%% file: main.tex
\documentclass{article} 

\input{preamble}

\begin{document}

\maketitle
\lhead{Preprint}
\input{sec/0_abstract}

\input{sec/1_intro}
\input{sec/2_related}
\input{sec/3_method}
\input{sec/5_experiments}

\input{sec/6_conclusion}

\bibliography{iclr2027_conference}
\bibliographystyle{iclr2027_conference}

\newpage
\appendix
\counterwithin*{figure}{section}
\counterwithin*{table}{section}
\renewcommand{\thefigure}{\thesection\arabic{figure}}
\renewcommand{\thetable}{\thesection\arabic{table}}
\input{sec/7_explanation_details}
\input{sec/7_experiments_details}
\input{sec/8_additional_results}

\input{sec/9_event_anotation}

\end{document}

%% file: preamble.tex
\usepackage{iclr2027_conference,times}
\iclrfinalcopy

\input{math_commands.tex}

\usepackage{hyperref}
\usepackage{url}
\usepackage{tcolorbox}
\usepackage{amssymb}
\usepackage{forest}
\tcbuselibrary{listings,breakable}

\title{Event-Aligned Visual Action Reasoning for World Action Models}

\author{Xiaomeng Yang\quad 
Yushu Wu\quad 
Yi Gao\quad 
Yuhao Lei\quad
Xuan Zhang\quad
Pu Zhao\quad
Yanzhi Wang \\
Northeastern University \\
\href{https://xiaomeng-yang.github.io/Event-aligned-WAM}{Project Page}
}

\usepackage{algorithm}
\usepackage{algpseudocode}
\usepackage{xspace}
\usepackage{amsmath}
\usepackage{multirow}
\usepackage{caption}
\usepackage{booktabs}
\usepackage{pifont}
\usepackage{graphicx}
\usepackage{tabularx}
\usepackage{wrapfig,lipsum}
\usepackage{subfig}
\usepackage[inline]{enumitem}
\usepackage{wrapfig}
\usepackage{threeparttable}

\usepackage[nameinlink]{cleveref}
\crefname{paragraph}{Sec.}{Secs.}
\Crefname{paragraph}{Sec.}{Secs.}
\crefname{section}{Sec.}{Secs.}
\crefname{table}{Tab.}{Tabs.}
\crefname{equation}{Eq.}{Eqs.}
\crefname{figure}{Fig.}{Figs.}

\newcommand{\onedot}{\ifx\@let@token.\else.\null\fi\xspace}

\newcommand{\vs}[0]{\emph{vs}\onedot}

\renewcommand{\paragraph}[1]{\noindent \textbf{#1.}}

\usepackage{listings}
\usepackage{xcolor}

%% file: math_commands.tex
\usepackage{amsmath,amsfonts,bm}

\def\eqref#1{equation~\ref{#1}}

\def\1{\bm{1}}

\DeclareMathAlphabet{\mathsfit}{\encodingdefault}{\sfdefault}{m}{sl}
\SetMathAlphabet{\mathsfit}{bold}{\encodingdefault}{\sfdefault}{bx}{n}



%% file: sec/0_abstract.tex
\begin{abstract}

World-Action Models (WAMs) utilize future visual prediction as an intermediate reasoning process to guide action generation. 
However, existing WAMs typically structure visual imagination according to predefined temporal intervals, without explicitly accounting for the different roles of task-critical interactions and connecting transitions. 
We argue that effective visual foresight should align directly with task-relevant interactions and their corresponding reasoning demands.
To this end, we introduce an event-aligned visual action reasoning framework that organizes visual-action prediction around interaction events.
Through event-aligned visual-action supervision, WAM learns to generate event-aligned visual context in each imagined rollout, placing greater emphasis on critical state changes that inform action generation. 
This shapes the visual reasoning granularity according to the underlying interaction dynamics, with detailed reasoning around task-critical events and coarser progression through connecting transitions. 
Furthermore, we introduce an execution validity head that identifies the valid portion of each predicted action sequence, avoiding redundant actions during chunked inference. 
Experiments demonstrate a 10.26 percentage point improvement in DOMINO success rate over baseline and competitive performance on RoboTwin 2.0.
It also transfers from DOMINO Level 1 to Levels 2 and 3 without target-level adaptation. 
\end{abstract}

%% file: sec/1_intro.tex
\section{Introduction}

Recent World Action Models~(WAMs) have emerged as a promising paradigm for robot manipulation by coupling action generation with future visual prediction \citep{li2026causalworldmodelingrobot,DBLP:journals/corr/abs-2606-02800,ye2026worldactionmodelszeroshot}.
Unlike conventional vision-language-action (VLA) policies that map current observations and instructions directly to actions \citep{pmlr-v270-kim25c,black2026pi0visionlanguageactionflowmodel,DBLP:journals/corr/abs-2503-14734,pmlr-v305-black25a,DBLP:journals/corr/abs-2503-20020,zheng2025xvlasoftpromptedtransformerscalable}, WAMs additionally model how the scene may evolve and leverage the predicted future for action generation.
Such visual prediction can be viewed as an intermediate reasoning process that simulates future scene evolution and grounds action generation in predicted visual states.
From this perspective, a fundamental question is how a WAM should structure its visual imagination to best inform action generation. 

A key observation is that visual states along a manipulation trajectory contribute differently to action generation.
Consider grasping an object: physical contact between the gripper and the object provides direct evidence of the interaction, while the preceding gripper closure and the subsequent lift provide the local context for how the grasp begins and proceeds.
Meanwhile, the initial reach serves as a connective phase linking earlier task stages to the grasp itself.
These phases serve distinct functions:
The core interaction (\texttt{grasp}) evidence confirms whether the intended robot–object state transition has occurred; the surrounding motion (\texttt{closing} and \texttt{lifting}) provides the context of this change; and the connecting motion (\texttt{approaching}) determines how the robot transitions to it.
However, the manipulation trajectory rarely reflects these functional task stages. 
Informative interaction events often span only a few frames, while connecting motions can take up the majority of the trajectory.
Therefore, visual imagination based on fixed temporal intervals may risk fragmenting critical, fine-grained interactions across distinct prediction horizons.

\input{fig/motivation}

Despite rapid progress in WAMs, current paradigms largely overlook this functional gap.
Existing works predominantly focus on how visual predictions should be integrated with actions.
\citet{li2026causalworldmodelingrobot} explicitly interleaves future visual prediction and action generation, \citet{yuan2026fastwamworldactionmodels} retains video modeling only during training, and \citet{zhang2026imagewamworldactionmodels} replaces dense future videos with task-relevant visual transformations.
However, as shown in ~\Cref{fig:motivation_a}, these methods universally structure visual-action horizons uniformly across the demonstration timeline (denoted as \textit{Episode Uniform}).
They largely inherit the sampled trajectory structure from training data, without adapting visual imagination to different interaction events.

To address the above limitations, we propose an event-aligned visual action reasoning framework that organizes visual imagination around task-critical interactions, rather than directly following the sampled trajectory structure. 
The key idea is to shape the temporal progression of imagined states according to the underlying interaction dynamics.
We construct event-aligned visual-action supervision by first segmenting training trajectories according to interaction events and then building visual-action prediction chunks from the resulting event boundaries.
Within this event-aligned organization, we reorganize training trajectories such that task-critical events preserve fine-grained visual-action evolution, while transitions toward the relevant states are represented with fewer intermediate states, as shown in~\Cref{fig:motivation}.
Through this supervision, the model learns to advance its visual imagination in alignment with interaction progress, generating visual context that captures how critical interactions unfold under actions and supports action generation.

However, event-aligned visual action reasoning also introduces a practical challenge for action execution.
Since interaction segments may span different durations, their corresponding action sequences can have different valid lengths.
When mapped to fixed-size action chunks following \citet{li2026causalworldmodelingrobot}, short segments use repeated or padded actions to fill the remaining prediction slots during training.
Executing these padding-induced actions can unnecessarily prolong
the segment before the next prediction.
To address this issue, we further introduce an execution validity head that identifies the valid portion of each predicted action sequence.
At inference time, the validity head determines which portion
to execute, avoiding redundant actions while retaining actual
actions associated with interactions.

Experiments on DOMINO~\citep{fang2026domino} and RoboTwin 2.0~\citep{chen2025robotwin} demonstrate improved manipulation performance, including a 10.26\% gain in DOMINO success rate over LingBot-VA~\citep{li2026causalworldmodelingrobot} and competitive performance on RoboTwin 2.0. 
The framework also demonstrates successful transfer from DOMINO Level 1 to Level 2/3 without target-level adaptation.
Our contributions are summarized as below,
\begin{itemize}[leftmargin=1.5em, itemsep=1pt, topsep=1pt, parsep=1pt, partopsep=1pt]
    \item We proposed an event-aligned visual action reasoning framework that organizes future prediction with task-critical interactions as focus to reason more carefully over crucial state changes.
    Instead of inheriting a predefined temporal structure from sampled trajectories, our formulation enables the reasoning process to adapt its granularity to the interaction dynamics.

    \item We further introduce an execution validity head to identify the valid portion of each predicted action sequence under event-aligned reasoning.
    By avoiding the execution of redundant actions, it improves action continuity and leads to smoother action execution.

    \item Our comprehensive experiments demonstrate that our method achieves superior performance, especially on the more challenging dynamic manipulation dataset DOMINO, with significantly better success rate and outstanding transfer performance.  

\end{itemize}

%% file: fig/motivation.tex

\begin{figure}[t]
    \centering
    \makebox[\linewidth][c]{%
        \subfloat[Visual-Action Chunks Granularity]{%
            \label{fig:motivation_a}%
            \includegraphics[width=0.70\linewidth]{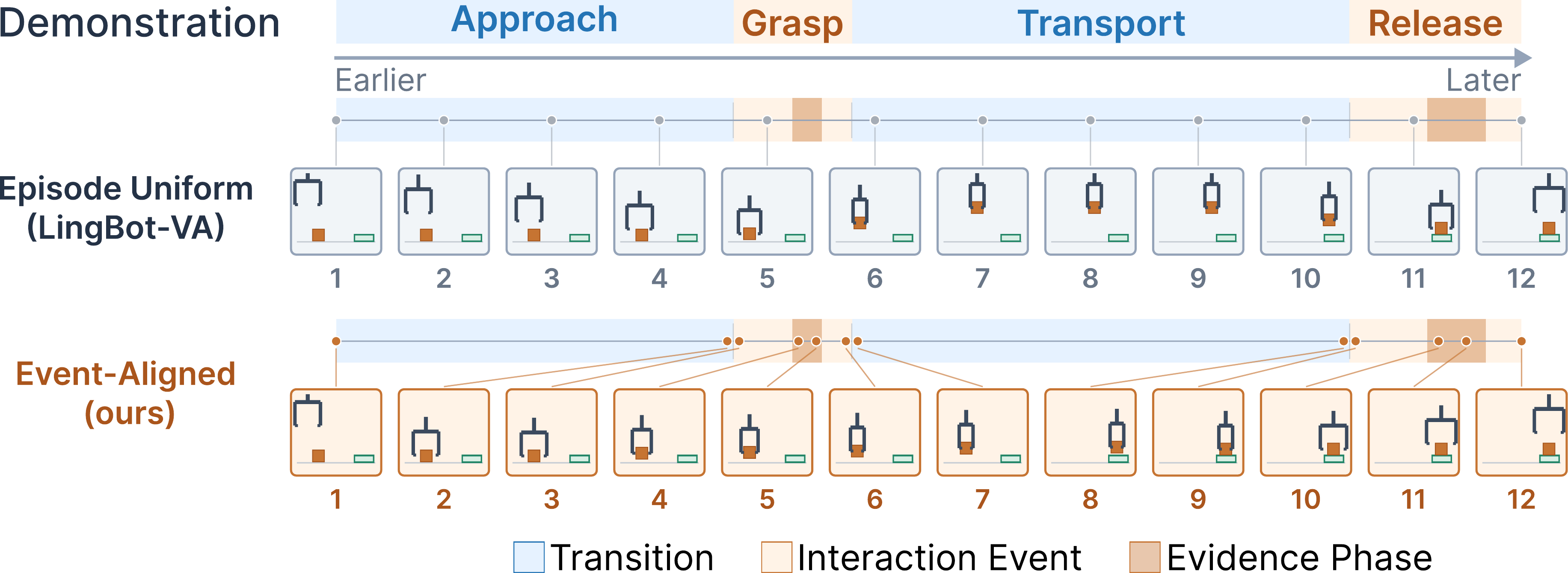}%
        }%
        \hfill%
        \subfloat[DOMINO success rate]{%
            \label{fig:motivation_b}%
            \includegraphics[width=0.28\linewidth]{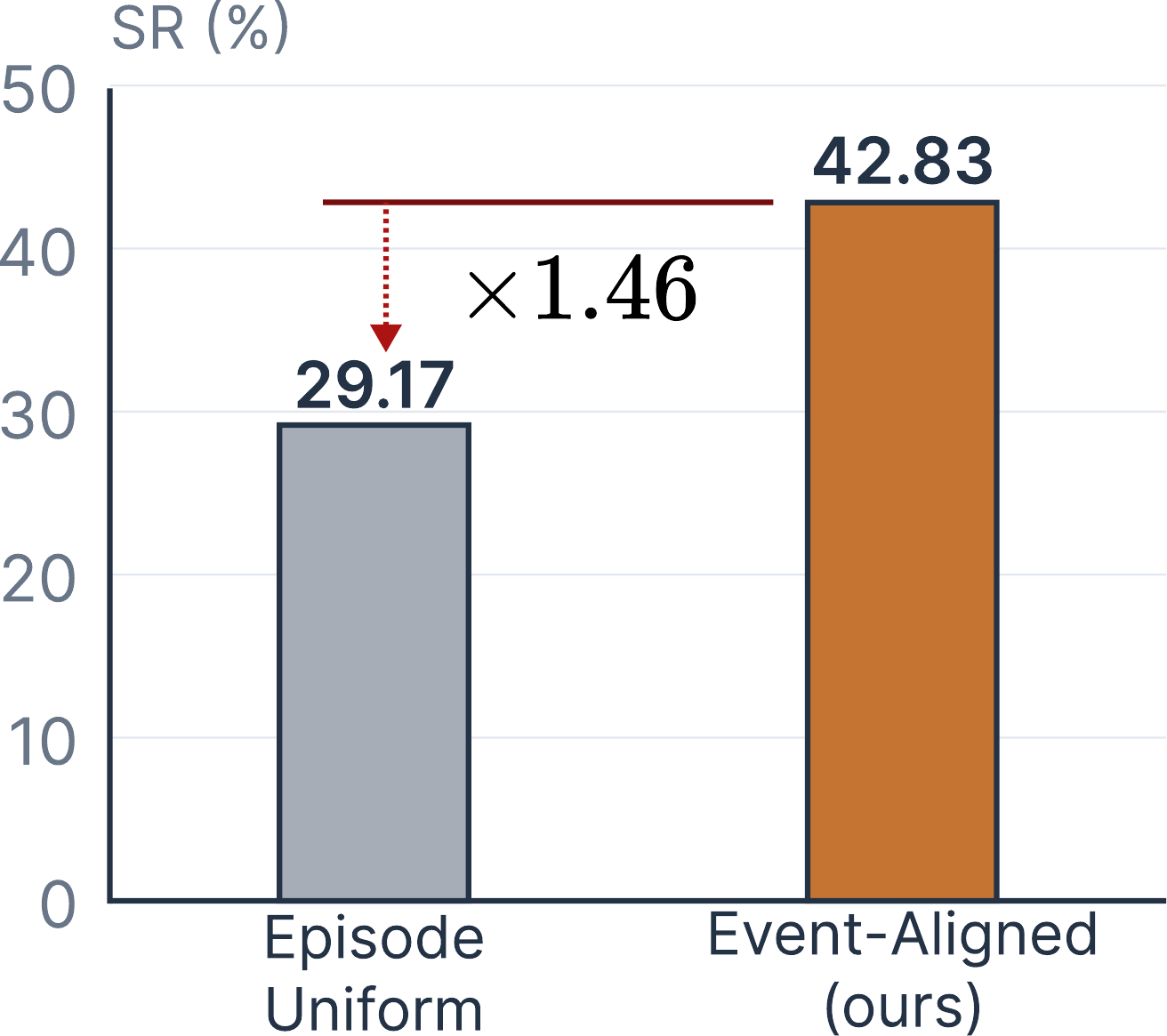}%
        }%
    }%
    \caption{
        \textbf{Interaction-dependent temporal granularity of visual reasoning.}
        (a) Our event-aligned visual action reasoning framework organizes
        visual-action targets around interaction events and transitions.
        (b) DOMINO success rates under a matched sequence budget of
        training visual-action slots.
    }
    \label{fig:motivation}
    \vspace{-20pt}
\end{figure}

%% file: sec/2_related.tex
\section{Related Works}

\paragraph{Vision-Language-Action Models}
VLA models transfer knowledge from pretrained VLMs to robot control by generating actions conditioned on vision observations and language instructions.
OpenVLA~\citep{pmlr-v270-kim25c} formulates control as action-token prediction, while $\pi_0$~\citep{black2026pi0visionlanguageactionflowmodel} and GR00T N1~\citep{DBLP:journals/corr/abs-2503-14734} combine pretrained VLM representations with generative action models.
Recent works further improve generalization across tasks and embodiments through heterogeneous data and scalable adaptation, as in $\pi_{0.5}$~\citep{pmlr-v305-black25a}, Gemini Robotics~\citep{DBLP:journals/corr/abs-2503-20020}, and X-VLA~\citep{zheng2025xvlasoftpromptedtransformerscalable}.
Other works investigate how pretrained representations should be adapted for action generation, including explicit spatial modeling \citet{qu2025spatialvlaexploringspatialrepresentations}, lightweight feature adaptation~\citet{10.1609/aaai.v40i22.38931}, representation insulation~\citet{NEURIPS2025_94e93603}, and persistent object-centric representations~\citet{ren2026closingloophumanoidvla}.
Despite their strong generalization, these approaches primarily map current observations and instructions directly to actions.
Recent methods such as MEM~\citep{torne2026memmultiscaleembodiedmemory} and LingBot-VLA 2.0~\citep{wu2026foundationapplicationimprovingvla} begin to move beyond this formulation by incorporating temporal memory and future prediction, respectively.

\paragraph{World-Action Models}
Beyond directly mapping observations to actions, recent  WAMs leverage video generative models to jointly model future visual dynamics and robot actions.
\citet{ye2026worldactionmodelszeroshot,li2026causalworldmodelingrobot,DBLP:journals/corr/abs-2606-02800,xu2026forcingcausalworldmodeling} couple visual prediction with action generation, using predicted future observations as an intermediate representation for reasoning. 
These approaches demonstrate that modeling the evolution of the visual world can provide richer structure than direct action prediction. 
Recent works further explore whether dense video prediction is necessary.
Fast-WAM~\citep{yuan2026fastwamworldactionmodels} reduces explicit visual generation at inference, while ImageWAM~\citep{zhang2026imagewamworldactionmodels} focuses on single target-image prediction instead of video rollout, suggesting that uniformly predicting every intermediate frame introduces substantial redundancy. 
In parallel, approaches explore temporal representation in world modeling.
\citet{li2026wallwmcarvingworldaction} models trajectories around action-relevant events, while \citet{he2026skipsparsekeyframeinterpolation,11246352} use sparse keyframes or subgoals to represent long-horizon dynamics.
These methods highlight the importance of selecting task-relevant temporal information rather than modeling all future states uniformly.

%% file: sec/3_method.tex
\section{Interaction-Structured Visual Reasoning for WAM}

We propose an event-aligned visual action reasoning framework that organizes visual prediction according to the reasoning demands of robot manipulation.
Rather than treating visual prediction as a uniform continuation of the sampled trajectory, we structure visual reasoning around task-relevant interactions and the visual context needed to support their corresponding actions.
Our framework consists of event-aligned visual-action reasoning and execution validity prediction.

\subsection{World-Action Models}
\label{sec:preliminary}
\paragraph{Problem Formulation}
We build on autoregressive latent-diffusion World-Action Models (WAMs)~\citep{li2026causalworldmodelingrobot}, which represents the episode
as a sequence of visual-action chunks $\{(\mathcal V_n,\mathcal A_n)\}_{n=1}^{N}$.
Visual observations are encoded by a causal video VAE~\citep{wan2025wan}, where each latent visual frame is temporally aligned with a group of $\tau$ actions, following  ~\citet{li2026causalworldmodelingrobot} sampling scheme. 

Given the initial visual context $\mathcal V_0$ encoded from $o_0$ and the instruction $l$, the model is trained to autoregressively generate the interleaved sequence
$\mathcal V_1,\mathcal A_1,\ldots,\mathcal V_N,\mathcal A_N$, producing each visual or action chunk through iterative denoising conditioned on its preceding context in the sequence.
It is optimized with teacher forcing, conditioning each prediction on ground-truth visual-action history.

At inference time, the model first predicts a visual chunk and then generates its associated action chunk conditioned on the imagined scene evolution.
Therefore, \textit{visual predictions act as a form of reasoning for action}.
The imagined scene provides the context where the action is derived.
As execution proceeds, the latest observations will replace their corresponding predicted visual states in the context history, grounding subsequent generation in the observed environment.

\input{fig/framework}

\paragraph{Trajectory-Based Visual Reasoning}
Under the standard WAM sampling scheme, visual-action prediction inherits the temporal organization of the demonstration episode.
For a chunk containing $K$ latent visual states,
\begin{equation}
\mathcal V_n=(z_{n,1},\ldots,z_{n,K}),\qquad
\mathcal A_n=(\mathbf a_{n,1_{\tau}},\ldots,\mathbf a_{n,K_{\tau}}),
\label{eq:visual_action_progression}
\end{equation}
where $z_{n,k}$ denotes the $k$-th latent visual state and
$\mathbf a_{n,k_{\tau}}$ its temporally aligned group of $\tau$ actions.
These visual-action pairs are determined by the sampled trajectory, and the chunks with fixed size are unaware of interaction boundaries. Consequently, a task-relevant interaction may be split into neighboring chunks, or a chunk may contain multiple scattered events. This information presented to the action model may lead to potential performance loss.
We therefore seek a structure that organizes visual reasoning according to task-relevant interactions rather than trajectory position alone.

\subsection{Event-Aligned Visual Reasoning}
\label{sec:event_aligned}
A manipulation task progresses through \emph{interactions} (such as grasping, releasing, or pressing), connected by \emph{transitions} (such as reaching and carrying), with different demands on visual prediction.
Interactions contain the state changes that decide task success and call for detailed reasoning, while transitions follow regular observation--action dynamics.
Inspired by human perception and motor control, which adapt processing to the demands of the current task state~\citep{todorov_optimal_2002,Zacks2007EventPA}, we organize visual reasoning structured by interaction.
We first define it as a taxonomy over the events of a demonstration, and then build visual action chunks from it.

\paragraph{Interaction Taxonomy}
A demonstration is represented as a sequence of contiguous events, where each event corresponds to a recognizable unit of behavior with a clear beginning and end. 
Formally, events $E_0, \ldots, E_M$ partition the episode, where the $i$-th event covers frames $s_i$ through $e_i$: 
\begin{equation}
    E_i=\bigl\{(o_t, a_t)\bigr\}_{t=s_i}^{e_i}, \quad e_M=T.
    \label{eq:events}
\end{equation}
Each event has a type $c_i\in\{\mathtt{interaction},\mathtt{transition}\}$.
An \emph{interaction event} changes a robot--object or robot--environment relationship that matters for task success.
A \emph{transition event} moves the robot between interactions and leaves these relationships unchanged.
Two interaction events may be adjacent with no transition in between them.

Within an interaction event, we further distinguish two phases as shown in \Cref{fig:interaction-taxonomy}.
The \emph{evidence phase} is the part in which the relationship change becomes visually observable, providing direct visual evidence of task progress.
The \emph{support phase} is the remaining motion within the same event that occurs before, after, or on both sides of the evidence phase.
The first and last frame of the evidence phase are its \emph{evidence boundaries} $s_i^\mathrm{ev}$ and $e_i^\mathrm{ev}$, with $s_i \leq s_i^\mathrm{ev} < e_i^\mathrm{ev}\leq e_i$.
The support phase is the complement $[s_i,s_i^\mathrm{ev})\bigcup(e_i^\mathrm{ev},e_i]$.
Transition events contain no task-progress evidence and are therefore not further divided.
\footnote{
In a grasp, the evidence phase is the gripper making contact with the object, the support phase is the gripper starting to close before contact and lifting off after it, and the reach before the grasp is a transition event.
}
Event and evidence boundaries come from the event annotation of the source demonstration and are fixed offline as shown in~\Cref{app:event-annotation}.

\input{algo/event_aligned_va}

\paragraph{Event-Aligned Interaction Structure}
We now build the visual-action chunk of \Cref{sec:preliminary} structured from events instead of fixed temporal positions.
Three rules connect the taxonomy to the chunks.
\begin{enumerate}[
    label=\textsc{Rule~\arabic*.},
    wide=0pt,
    leftmargin=0pt,
    topsep=0pt,
    itemsep=0pt
]
    \item \emph{Event Boundaries}:
    A chunk never crosses an event boundary.
    Each transition event is resampled into exactly one chunk, while an interaction event produces one or more chunks.

    \item \emph{Phase-dependent density}: 
    Within an interaction event, the evidence phase keeps its native temporal resolution and is never subsampled.
    The support phase is resampled into slots that the evidence leaves free in its chunks.
    If the evidence leaves no slot, the support before or after it forms a chunk of its own.

    \item \emph{Interaction Boundaries}:
    The first and last frame of an interaction event are always kept.
    Where a support phase lies on the side, its resampling anchors the boundary frame.
\end{enumerate}
Therefore, both levels of the taxonomy enter the construction.
Event boundaries decide where a chunk may start and end, and phases decide which frames inside an interaction are kept densely.
Support can be compressed while keeping its boundaries, indicating how the event begins and ends around the evidence.
\Cref{sec:exp-analysis} shows that reducing it hurts even when evidence itself is protected.
\Cref{alg:event-aligned-chunks} shows the construction algorithm.
It uses three operators: $\mathcal S_r(\cdot)$ resamples a segment to $r$ frames, $\bar{\mathcal S}_r(\cdot)$ does the same while keeping the first and last frame of the segment, and $\mathrm{Split}_K(\cdot)$ cuts a sequence into consecutive chunks of $K$ frames.
Each action chunk consists of the actions paired with the selected frames.
If a segment is shorter than its allocated slot count, frames are repeated as needed.
These repeated frames carry no new observation and are exactly the positions that the execution validity head of \Cref{sec:validity_head} learns to skip.
More details are discuss in~\Cref{app:construction-detail}.

Aligning chunk boundaries with events also matters at inference time.
Because a chunk never mixes two events, an interaction that does not complete as expected is revisited by the next prediction rather than averaged into a chunk that already contains the following motion, so prediction drift is corrected where it arises instead of propagating across interactions.

\subsection{Execution Validity Prediction}
\label{sec:validity_head}
Event-aligned chunks have a fixed length $K$, but the segments they encode do not.
A segment shorter than its slots is padded by repeating frames before VAE encoding~(\Cref{sec:event_aligned}), and each repeated frame carries a copy of the action group of the frame it repeats.
At inference, the event boundaries are unknown, so the model can not know how many of the $K$ slots of a chunk carry new observations.
Executing the copies would make the robot pause at the same target, which we call a \emph{stall}.
Therefore, for every generated action group, we let the model predict whether it should be executed.

\paragraph{Joint Action and Validity Prediction}
Let $h_n$ denote the context available before predicting chunk $n$, including the instruction $l$, initial context $\mathcal V_0$, and all previous visual-action chunks.
Conditioned on predicted visual chunk $\widehat{\mathcal V}_n$ and $h_n$, the model generates  action chunk $\widehat{\mathcal A}_n=(\widehat a_{n,1},\ldots,\widehat a_{n,K_{\tau}})$ together with a validity probabilities $p_n=(p_{n,1}, \ldots, p_{n,K})$,
\begin{equation}
    \bigl(\widehat{\mathcal A}_n,\widehat{\mathbf p}_n\bigr)
    = f_\theta\bigl(\widehat{\mathcal V}_n,h_n\bigr),
    \label{eq:validity_prediction}
\end{equation}
where $\widehat{\mathbf p}_n=(\widehat p_{n,1},\ldots,\widehat p_{n,K_{\tau}})$ denotes the probabilities that each action should be executed.
The probabilities are predicted by a projection layer $g_\phi$ applied to the action-stream features of each of the $K$ positions after the final transformer block, which outputs a logit $\ell_{n_k}$ with $p_{n,k}=\sigma(\ell_{n,k})$.
The head is trained jointly with the backbone.
The ground-truth mask $\mathbf m_n$ comes from chunk construction, where $m_{n,k}=1$ if the $k$-th frame of action is the first occurrence of an observation frame, and $m_{n,k}=0$ if it is a repeated copy.
Validity is defined per frame and shared by the $\tau$ actions aligned with it.
The head is supervised with cross-entropy over all chunks
and positions, $\mathcal L_{\mathrm{val}}
=-\tfrac{1}{K\tau}\sum_{j=1}^{K\tau}
[m_{n,j}\log p_{n,j}+(1-m_{n,j})\log(1-p_{n,j})]$.

\paragraph{Stall-Free Execution}
At inference time, we set $\widehat m_{n,j}=\mathbf{1}[\sigma(\ell_{n,j})>0.5]$ and execute only the valid action groups, in their original order,
\begin{equation}
    \widehat{\mathcal A}^{\mathrm{exec}}_n
    =\bigl(\widehat a_{n,j}\bigr)_{\substack{1\le j\le K,\ \widehat m_{n,j}=1}}.
    \label{eq:stall_free_execution}
\end{equation}
This mechanism allows the executed action sequence to adapt to the interaction represented by each chunk while the model keeps a fixed-length output structure.

\subsection{Training Objective}
\label{sec:objective}
Following~\citet {li2026causalworldmodelingrobot}, we train the visual and action streams on the event-aligned chunks using flow-matching losses $\mathcal L_{\mathrm{dyn}}$ and $\mathcal L_{\mathrm{inv}}$ for visual and action prediction, respectively.
$\mathcal L_{\mathrm{dyn}}$ is applied to all frames, including repeated frames, so the visual stream learns to reproduce the padded chunk structure.
$\mathcal L_{\mathrm{inv}}$ is restricted and averaged only over valid positions, and invalid action loss is zeroed out.
The overall objective is
$\mathcal L=\mathcal L_{\mathrm{dyn}}
+\mathcal L_{\mathrm{inv}}+0.1\mathcal L_{\mathrm{val}}$.

%% file: fig/framework.tex
\begin{figure}[t]
    \centering
    \subfloat[
    \textbf{Event-aligned visual-action chunk construction and joint training.}
    ]{\includegraphics[width=0.95\linewidth]{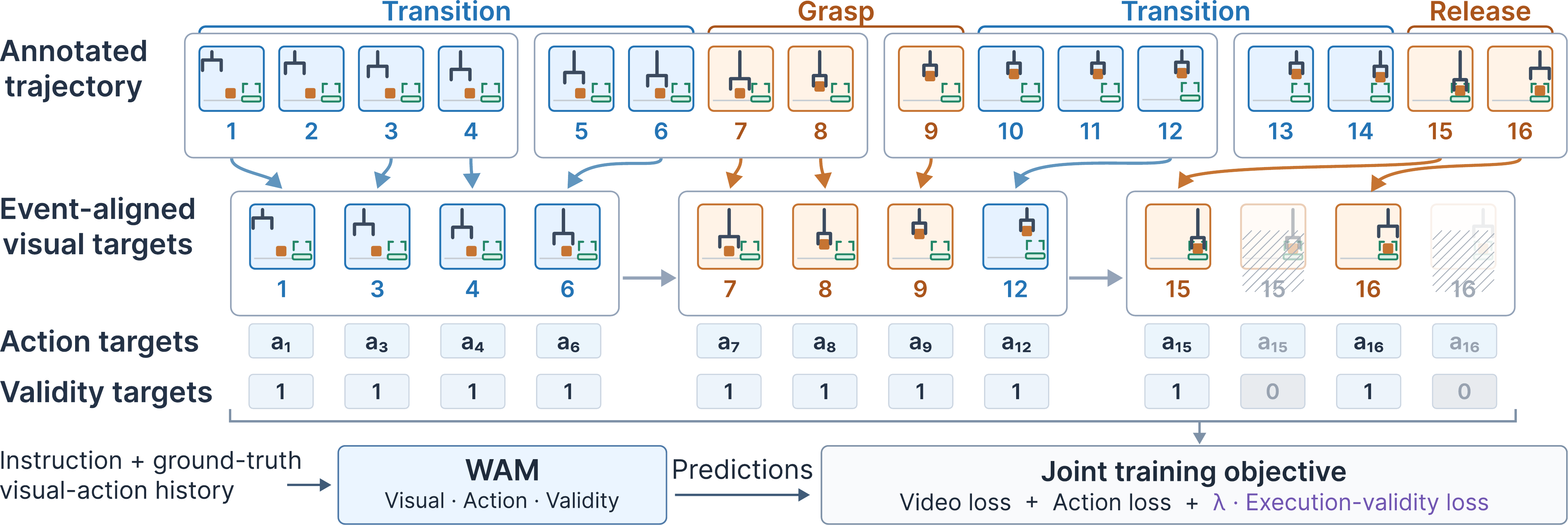}}
    \vfill
    \subfloat[
    \textbf{Visual-guided action generation with execution validity prediction.}
    ]{\includegraphics[width=0.95\linewidth]{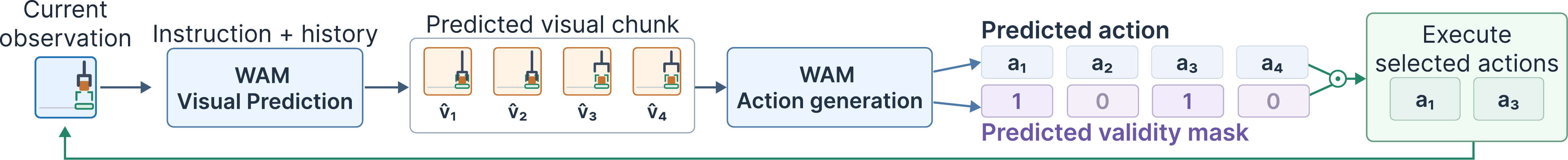}}
    \caption{
    \textbf{Interaction-Structured Visual Reasoning framework.}
    Event-aligned chunks allocate visual reasoning according to interaction structure, while execution validity prediction identifies which generated actions to execute.
    }
    \label{fig:framework}
\end{figure}

%% file: algo/event_aligned_va.tex
\begin{algorithm}[t]
\caption{Event-aligned visual-action chunk construction}
\label{alg:event-aligned-chunks}
\begin{algorithmic}[1]
\Require Events $\{(E_i,c_i)\}_{i=0}^{M}$ with evidence boundaries of interaction events; chunk length $K$
\Ensure Sequence of visual-action chunks $\mathcal C$
\State $\mathcal C\gets[\,]$
\For{$i=0,\ldots,M$}
    \If{$c_i=\mathtt{transition}$}
        \State $\mathcal C.\mathrm{append}(\mathcal S_K(E_i))$ \Comment{One chunk per transition, endpoints not anchored}
    \Else
        \State $P,V,Q\gets$ leading support, evidence, trailing support of $E_i$
        \State $F\gets\lceil |V|/K\rceil \cdot K-|V|$ \Comment{Slots the evidence leaves free}
        \If{$F=0$} \Comment{Evidence fills whole chunks; Support phases standalone}
            \If{$P\neq\emptyset$} 
                \State $\mathcal C.\mathrm{append}(\bar{\mathcal S}_K(P))$
            \EndIf
            \State $\mathcal C.\mathrm{extend}(\mathrm{Split}_K(V))$
            \If{$Q\neq\emptyset$}
            \State $\mathcal C.\mathrm{append}(\bar{\mathcal S}_K(Q))$
            \EndIf
        \Else \Comment{Support fills the free slots around the evidence}
            \State $(n_P,n_Q)\gets\mathrm{Allocate}(F,P,Q)$ \Comment{$\ge 1$ slot per non-empty side, rest split evenly}
            \State $\mathcal C.\mathrm{extend}\bigl(\mathrm{Split}_K(\bar{\mathcal S}_{n_P}(P)\oplus V\oplus\bar{\mathcal S}_{n_Q}(Q))\bigr)$
        \EndIf
    \EndIf
\EndFor
\State \Return $\mathcal C$
\end{algorithmic}
\end{algorithm}

%% file: sec/5_experiments.tex
\providecommand{\expblank}{\textemdash}
\providecommand{\expfield}{\underline{\hspace{1.0cm}}}

\section{Experiments}
\label{sec:experiments}

\subsection{Experimental setup}
\label{sec:exp-setup}

\paragraph{Benchmarks and baselines}
We evaluate on DOMINO~\citep{fang2026domino} and RoboTwin~2.0~\citep{chen2025robotwin}. DOMINO contains 35 dynamic-manipulation tasks. 
We use clean Level~1 scenes with the Aloha-AgileX embodiment and a motion coefficient of 0.1, evaluating 100 episodes per task. 
RoboTwin~2.0 contains 50 bimanual-manipulation tasks, which we evaluate separately in clean and randomized environments. 
We compare against LingBot-VA~\citep{li2026causalworldmodelingrobot}, Fast-WAM~\citep{yuan2026fastwamworldactionmodels}, and ImageWAM~\citep{zhang2026imagewamworldactionmodels}.
We retain each baseline's original architecture and preprocessing pipeline and train all baselines for a comparable number of epochs. We additionally evaluate our framework on real-world manipulation tasks, with the experimental setup and results provided in \Cref{sec:exp-realworld}.

\paragraph{Implementation and metrics}
Our model is initialized from pretrained LingBot-VA~\citep{li2026causalworldmodelingrobot}.
We report Success Rate (SR) and Manipulation Score (MS) on DOMINO, and SR under clean and random settings on RoboTwin 2.0. 
MS measures terminal spatial progress toward the target with penalties for workspace violations and clutter collisions.
We implement the execution head as a zero-initialized token-wise linear projection, with an execution-loss weight of $\lambda_{\mathrm{exec}}=0.1$ and an inference threshold of $\tau_{\mathrm{exec}}=0.5$. 
Dataset preparation, model-specific hyperparameters, and evaluation details are provided in ~\cref{app:experiment-protocol}.

\subsection{Benchmark Performance}
\label{sec:exp-main-results}

\input{table/main_table}

\input{table/ablation_generalization}

\paragraph{DOMINO}
As shown in \Cref{tab:combined-benchmarks}, on DOMINO, our method achieves 42.83\% SR and 55.92 MS, outperforming the LingBot-VA backbone (32.57\% SR and 45.55 MS) with non-marginal improvements of 10.26 in SR and 10.37 in MS.
Note that although Fast-WAM and ImageWAM can achieve above 90\% SR on RoboTwin 2.0, their SR on DOMINO are below 20\%, 
highlighting the challenges of dynamic manipulation on DOMINO.
Our method can significantly improve performance on dynamic manipulation, demonstrating the effectiveness of event-aligned visual reasoning. Complete task-level results are provided in \Cref{app:per-task-results}.

\paragraph{RoboTwin 2.0}
As shown in \Cref{tab:combined-benchmarks}, on RoboTwin~2.0, our method obtains 93.02\% SR in clean scenes and 91.58\% in randomized scenes, with an average of 92.30\%, outperforming the  corresponding LingBot-VA backbone. Compared with ImageWAM and Fast-WAM, our method leads  to competitive performance, while performing much better than $\pi_{0.5}$   and ABot-M0. 

\paragraph{Transfer across DOMINO levels}
We further evaluate the transfer performance where the model is trained on the Level~1 data of DOMINO and evaluated on the  ten-task Level~2/3 subset without target-level additional adaptation.
As shown in \Cref{tab:ood}, our method achieves the best SR and MS performance with non-marginal improvements, such as our 45.42 MS \vs 31.56 MS from Fast-WAM on L2, and our 24.23 MS \vs 20.16 MS from PUMA on L3.
Note that our zero-shot model performs much better than PUMA, which finetunes the L1 model with LoRA for L2/L3, demonstrating our outstanding generalization and robustness performance.
The detailed results are shown in \Cref{app:per-task-results}.

\subsection{Analysis of event-aligned visual action reasoning}
\label{sec:exp-analysis}

We examine how the temporal organization of supervision affects event-aligned visual action reasoning, considering \emph{visual reasoning granularity}, the \emph{placement of interaction evidence}, and the \emph{surrounding context}.
Our Event-aligned sampling serves as the reference scheme: it organizes demonstrations into event-aligned granularity, explicitly preserves interaction evidence, and retains slots from the surrounding non-evidence context.  
The sampling variants in \Cref{tab:main-ablation} use the same 1,693 retained DOMINO demonstrations and share the same training and execution head configuration.
We additionally assess the execution validity head through a separate comparison of models trained with and without the head.
LingBot-VA without the head is included as a backbone reference.
The composition of sampled sequences relative to evidence phases is detailed in~\Cref{app:sequence-composition}. 

\input{table/main_ablation}

\paragraph{Event alignment at a matched sequence budget}
Episode Uniform samples uniformly across each demonstration’s initial-to-final range, preserving the endpoints and total action-slot budget of Event-aligned. 
Event Uniform instead preserves its event boundaries and per-event budgets, while sampling uniformly within each event without explicitly reserving interaction evidence. 
Event Uniform reaches 39.60\% SR, much higher than 29.17\% from Episode Uniform, under exactly the same slot budget. It demonstrates that event-aligned granularity and their associated budget allocation provide a more effective temporal organization than episode-wide uniform sampling. 

\paragraph{Interaction evidence placement}
Explicitly preserving interaction evidence further improves SR from 39.60\% with Event Uniform to 42.83\% with Event-aligned, while maintaining the same event boundaries and per-event budgets. 
To further examine the role of evidence placement, Reduced Evidence retains the event boundaries but shifts sampling toward transition event and support phase, away from interaction evidence. 
Despite using 268,592 action slots (26.9\% more than Event-aligned), it achieves only 36.17\% SR, demonstrating the importance of preserving fine-grained interaction evidence.

\paragraph{Context surrounding protected interaction evidence}
Reduced Context preserves the same protected interaction evidence as Event-aligned while decreasing the sampling budget for surrounding non-evidence context, resulting in less action slots.
It reaches 37.83\% SR, lower than Event-aligned, indicating the effectiveness of additional context around protected interaction evidence. Our framework benefits from both evidence and surrounding support. 

\paragraph{Execution validity prediction}
We assess the execution validity head under Event-aligned sampling by comparing separately trained models with and without the head.
The model without the head directly executes the complete predicted action sequence.
As shown in ~\Cref{tab:main-ablation}, the model with the head improves SR from 39.11\% to 42.83\%, showing the effectiveness of distinguishing action validity with the proposed execution validity head.

\input{fig/attention}
\paragraph{Visualization of event-aligned visual reasoning}
\Cref{fig:attention_viz} compares LingBot-VA and our method on a \textit{Scan Object} episode, exhibiting video attention to the previous observation (\Cref{fig:attention_viz}a), and action attention to the visual prediction (\Cref{fig:attention_viz}c). 
Both methods start from the same initial observation, and later chunks follow each policy's own rollout.
At chunk 0, our method's video attention already focuses on the moving object, while LingBot-VA's is largely diffuse.
Our predicted visual futures in \Cref{fig:attention_viz}c depict an ordered interaction sequence:
approaching and grasping the moving object (chunks~0--1), 
approaching and grasping the scanner (chunks~2--3),
and reaching a scanning configuration (chunk~5).
Action attention shifts toward the corresponding interaction regions, highlighted by the red boxes.
In contrast, LingBot-VA splits its attention across both arms and attempts to grasp both objects simultaneously, missing the moving target.
Our method completes the task at observation~140, while LingBot-VA remains unsuccessful at observation~496 (\Cref{fig:attention_viz}b).
These observations are consistent with our finding that task-critical events demand detailed reasoning and suggest that organizing visual prediction around interaction events benefits task completion.
Additional visualization cases are given in~\Cref{app:qualitative}.

\subsection{Execution behavior}

We examine the number of action commands and visual-rollout requests under our policy and the original LingBot-VA, on identical episodes where both can succeed. 
As demonstrated in \Cref{tab:execution-efficiency}, across 738 identical successful episodes spanning all 35 tasks, the average issued command count decreases from 167.18 to 106.46, while the average visual-rollout request count drops from 6.22 to 4.49.
These results together show our method is not only more effective with higher SR, but also more efficient with less commands and visual predictions.
Task-level breakdowns and the scope of this comparison are given in \Cref{app:execution-breakdown}.

%% file: table/main_table.tex
\begin{table}[t]
\centering
\small
\caption{
\textbf{Evaluation on DOMINO and RoboTwin 2.0.}
For LingBot-VA on RoboTwin, the upper row reports published results, while the lower row ($\dagger$) reports our re-evaluation of the released checkpoint (3 steps for video tokens to $s=0.6$, 10 steps for action tokens to $s=1.0$).}
\label{tab:combined-benchmarks}
\small
\begin{tabular}{lcccccc}
\toprule
\multirow{2}{*}{\textbf{Methods}} & \multicolumn{2}{c}{\textbf{DOMINO}} & & \multicolumn{3}{c}{\textbf{RoboTwin 2.0}} \\
\cmidrule{2-3} \cmidrule{5-7}
& \textbf{SR}$\uparrow$ & \textbf{MS}$\uparrow$ & & \textbf{Clean SR}$\uparrow$ & \textbf{Rand. SR}$\uparrow$ & \textbf{Avg. SR}$\uparrow$ \\
\midrule
$\pi_0$~\citep{black2026pi0visionlanguageactionflowmodel} & 8.17 & 23.96 & & 65.92 & 58.40 & 62.16 \\
$\pi_{0.5}$~\citep{pmlr-v305-black25a} & 9.63 & 26.17 & & 82.74 & 76.76 & 79.75 \\
PUMA~\citep{fang2026domino} & 17.20 & 34.97 & & -- & -- & -- \\
DynamicWAM\footnotemark[\value{footnote}]~\citep{lou2026dynamicwamdualpathmotionconditioning}
& 38.20 & 53.20 & & -- & -- & -- \\
ABot-M0~\citep{yang2026abot} & -- & -- & & 81.20 & 80.40 & 80.80 \\
Motus~\citep{bi2026motus} & -- & -- & & 88.66 & 87.02 & 87.80 \\
\midrule
ImageWAM~\citep{zhang2026imagewamworldactionmodels} & 18.86 & 35.47 & & \textbf{93.20} & \textbf{93.56} & \textbf{93.38} \\
Fast-WAM~\citep{yuan2026fastwamworldactionmodels} & 19.09 & 35.33 & & 91.88 & 91.78 & 91.83 \\
\multirow{2}{*}{LingBot-VA$^\dagger$~\citep{li2026causalworldmodelingrobot}} & \multirow{2}{*}{32.57} & \multirow{2}{*}{45.55} & & 92.93 & 91.55 & 92.24 \\
 & & & & 92.18\rlap{$^\dagger$} & 90.24\rlap{$^\dagger$} & 91.21\rlap{$^\dagger$} \\
\midrule
\textbf{Ours} & \textbf{42.83} & \textbf{55.92} & & 93.02 & 91.58 & 92.30 \\
\bottomrule
\end{tabular}
\end{table}
\footnotetext{DynamicWAM is trained with 150 clean and 150 randomized demonstrations per task (10{,}500 in total).}

%% file: table/ablation_generalization.tex
\begin{table}[t]
\begin{minipage}[t]{0.48\linewidth}
\small
\centering
\vspace{0pt}
\captionof{table}{Cross-level transfer on the 10-task subset of DOMINO.}
\label{tab:ood}
\scalebox{0.95}{%
\begin{tabular}{lcccc}
\toprule
\multirow{2}{*}{\textbf{Method}} & \multicolumn{2}{c}{\textbf{L2 transfer}} & \multicolumn{2}{c}{\textbf{L3 transfer}} \\
\cmidrule(lr){2-3}\cmidrule(lr){4-5}
& \textbf{SR}$\uparrow$ & \textbf{MS}$\uparrow$ & \textbf{SR}$\uparrow$ & \textbf{MS}$\uparrow$ \\
\midrule
PUMA & 10.5 & 30.28 & 4.6 & 20.16 \\
ImageWAM & 13.3 & 30.44 & 2.8 & 15.09 \\
Fast-WAM & 13.5 & 31.56 & 2.4 & 15.34 \\
LingBot-VA & 16.9 & 30.37 & 6.8 & 19.36 \\
Ours & \textbf{29.4} & \textbf{45.42} & \textbf{9.7} & \textbf{24.23} \\
\bottomrule
\end{tabular}}
\end{minipage}
\hfill
\begin{minipage}[t]{0.48\linewidth}
\small
\centering
\vspace{0pt}
\captionof{table}{
\textbf{Execution behavior on DOMINO.} 
$N_{\rm exec}$ and $N_{\rm roll}$ are the average number of action commands issued and visual rollouts requested per episode, measured only on episodes where both LingBot-VA and our model succeed.}
\label{tab:execution-efficiency}
\scalebox{1}{%
\begin{tabular}{@{}lrrr@{}}
\toprule
\textbf{Method} & \textbf{SR}$\uparrow$ & {\boldmath $N_{\rm exec}$} $\downarrow$ & {\boldmath $N_{\rm roll}$} $\downarrow$ \\
\midrule
LingBot-VA & 32.57 & 167.18 & 6.22 \\
Ours & 42.83 & 106.46 & 4.49 \\
\bottomrule
\end{tabular}}
    \end{minipage}
\end{table}

%% file: table/main_ablation.tex
\begin{table}[t]
\centering
\small
\caption{
\textbf{Ablations of temporal sampling scheme and execution validity prediction on DOMINO.}
Action slots count all sampled positions, including repetitions, across all retained demonstrations.
All variants except the backbone reference and the no-head ablation use the execution validity head during training and inference. Full sampling statistics are reported in~\Cref{tab:sampling-statistics}.}
\label{tab:main-ablation}
\begin{tabular}{lllrrr}
\toprule
\textbf{Sampling} & \textbf{Granularity} & \textbf{Placement} & \textbf{Slots}
    & \textbf{SR}$\uparrow$ & \textbf{MS}$\uparrow$ \\
\midrule
LingBot-VA (reference)
    & Episode & Uniform
    & 291,296 & 32.57 & 45.55 \\
\midrule
Episode Uniform
    & Episode & Uniform
    & 211,664 & 29.17 & 41.32 \\
Event Uniform
    & Event & Uniform
    & 211,664 & 39.60 & 52.26 \\
\midrule
Reduced Evidence
    & Event & Transition-Protected
    & 268,592 & 36.17 & 48.36 \\
Reduced Context
    & Event & Evidence-protected
    & 173,216 & 37.83 & 52.99 \\
\textbf{Event-aligned (Ours)} 
    & Event & Evidence-protected
    & 211,664 & \textbf{42.83} & \textbf{55.92} \\
\midrule
\midrule
\multicolumn{3}{l}{Event-aligned w/o execution validity head}
    & 211,664 & 39.11 & 52.33 \\
\multicolumn{3}{l}{\textbf{Event-aligned w/ execution validity head}} & 211,664 & \textbf{42.83} & \textbf{55.92}\\
\bottomrule
\end{tabular}
\end{table}

%% file: fig/attention.tex
\begin{figure}
    \centering
    \includegraphics[width=0.8\linewidth]{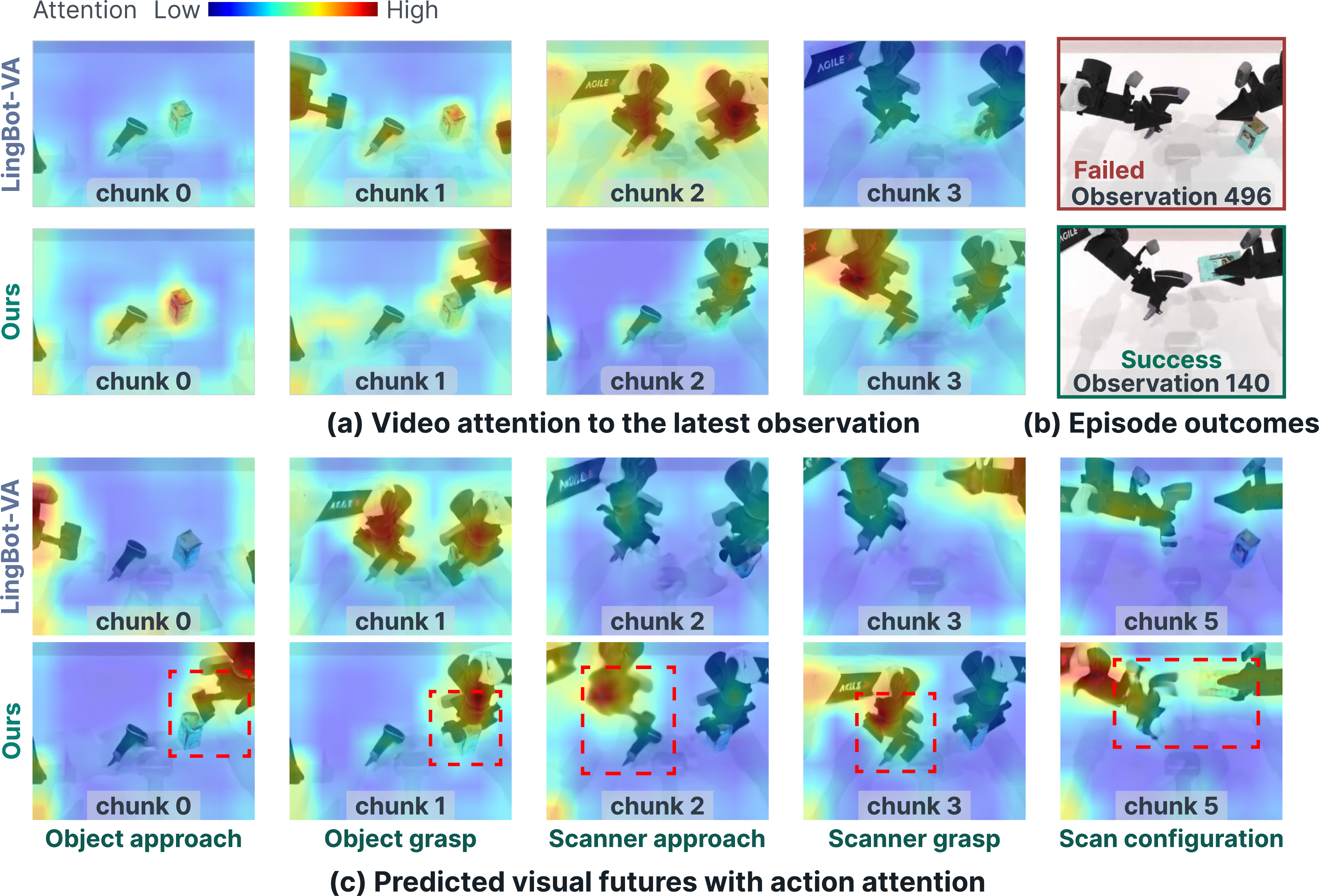}
    \caption{
    \textbf{Visual predictions and attention on a \textit{Scan Object} demonstration episode.} 
    (a) Video attention to the latest observation.
    (b) Episode outcomes.
    (c) The last predicted frame of each displayed chunk, overlaid with action attention. 
    Red boxes highlight interaction regions.
    Both methods start from the same initial observation. 
    Later chunks follow each policy's own rollout, so frames in the same column can differ.
    }
    \label{fig:attention_viz}
\end{figure}

%% file: sec/6_conclusion.tex
\section{Conclusion}

We presented an event-aligned visual action reasoning framework that structures visual imagination around task-relevant interactions and their corresponding reasoning demands. 
Through event-aligned visual-action supervision, our model learns to generate visual context for action generation with detailed state evolution around interaction events and coarser progression through connecting transitions, adapting visual reasoning granularity to the underlying interaction dynamics. 
An execution validity head identifies the valid portion of predicted action sequences to avoid redundant commands.
Experiments demonstrate state-of-the-art success rate on DOMINO and competitive performance on RoboTwin.
On shared successful episodes, it also reduces the average number of issued action commands and imagination requests compared with the baseline.
These findings highlight interaction structure as an effective organizing principle for visual imagination in WAMs.

%% file: sec/7_explanation_details.tex
\section{Details of Event-Aligned Chunk Construction}
\label{app:construction-detail}
\label{app:chunk-construction}

This section completes \Cref{alg:event-aligned-chunks} with the definitions and cases omitted from \Cref{sec:event_aligned}.
Throughout, a segment is a contiguous run of frames, each paired with its group of $\tau$ actions (\Cref{sec:preliminary}), and $K$ is the chunk length.

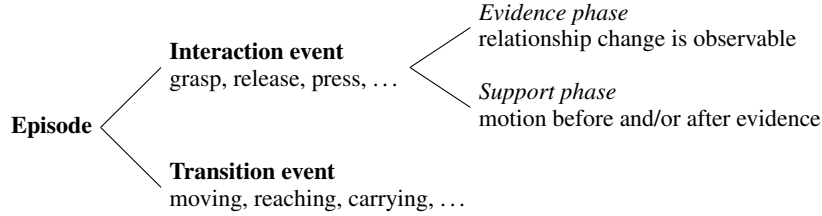
\begin{figure}[h]
\centering
\begin{forest}
for tree={
    grow'=0,
    parent anchor=east,
    child anchor=west,
    anchor=west,
    edge={thin},
    font=\small,
    l sep=8mm,
    s sep=2mm,
    inner xsep=3pt,
    inner ysep=2pt,
    align=left,
}
[{\textbf{Episode}}
    [{\textbf{Interaction event}\\
      \footnotesize grasp, release, press, \ldots}
        [{\emph{Evidence phase}\\
          \footnotesize relationship change is observable}]
        [{\emph{Support phase}\\
          \footnotesize motion before and/or after evidence}]
    ]
    [{\textbf{Transition event}\\
      \footnotesize moving, reaching, carrying, \ldots}]
]
\end{forest}
\caption{Interaction taxonomy of a demonstration. An episode is
partitioned into interaction and transition events. Interaction events are further structured into evidence and support phases.}
\label{fig:interaction-taxonomy}
\end{figure}

\paragraph{Taxonomy}
\Cref{fig:interaction-taxonomy} shows how we break a demonstration into parts, with a few examples at each level.
At the top is the episode, which we split into a sequence of events. 
In an \emph{interaction event}, the robot changes its relationship with an object or with the environment.
A \emph{transition event} covers the motion in between.
We split each interaction event further into two phases.
The \emph{evidence phase} consists of the frames that show the change happening.
The \emph{support phase} is the motion leading into or out of the evidence. 

\paragraph{Operators}
For a segment $X=(x_1,\ldots,x_{|X|})$ and a target length $r\ge 1$, the resampler $\mathcal S_r(X)$ returns $r$ frames taken at uniformly spaced positions of $X$, so that frames repeat when $|X|<r$; $\mathcal S_0(X)=\emptyset$.
The anchored resampler $\bar{\mathcal S}_r(X)$ keeps $x_1$ and $x_{|X|}$ and fills the remaining $r-2$ positions with $\mathcal S_{r-2}$ applied to the interior $(x_2,\ldots,x_{|X|-1})$; for $r=1$ it keeps the frame at the event boundary, so that Rule~3 holds.
$\mathrm{Split}_K(X)$ cuts $X$ into consecutive chunks of $K$ frames; if $|X|$ is not a multiple of $K$, the final chunk is filled to $K$ frames by repetition.
Repeated frames carry no new observation; they are the positions that the execution validity head (\Cref{sec:validity_head}) marks as invalid.

\paragraph{Transition events}
A transition event $E_i$ becomes the single chunk $\mathcal S_K(E_i)$.
Its first and last frame are not anchored.

\paragraph{Interaction events}
Let $P$, $V$, and $Q$ be the leading support, the evidence, and the trailing support of an interaction event, with $|V|\ge 1$ and either support possibly empty.
The evidence keeps every frame and occupies $\lceil |V|/K\rceil$ chunks, leaving $F=\lceil |V|/K\rceil K-|V|\in\{0,\ldots,K-1\}$ free slots.

\emph{Case $F>0$.}
The free slots are given to the support, $(n_P,n_Q)=\mathrm{Allocate}(F,P,Q)$ with $n_P+n_Q=F$:
if only one support is non-empty, it receives all $F$ slots;
if both are non-empty, each receives one slot and the remaining $F-2$ are split evenly, $(n_P,n_Q)=(\lfloor F/2\rfloor,\,F-\lfloor F/2\rfloor)$ for $F\ge 2$;
if neither exists, $(n_P,n_Q)=(0,0)$ and $\mathrm{Split}_K$ fills the free slots by repetition.
The sequence $\bar{\mathcal S}_{n_P}(P)\oplus V\oplus\bar{\mathcal S}_{n_Q}(Q)$ then has exactly $\lceil |V|/K\rceil K$ frames and is cut by $\mathrm{Split}_K$.

\emph{Case $F=0$.}
The evidence fills its chunks exactly.
Each non-empty support becomes a standalone chunk, $\bar{\mathcal S}_K(P)$ placed before the evidence chunks and $\bar{\mathcal S}_K(Q)$ after them.

\paragraph{Worked example}
Let $K=4$ and consider a grasp with $|P|=6$, $|V|=10$, and $|Q|=4$, preceded by a reach of $30$ frames.
The reach is a transition event and becomes one chunk of four frames sampled from its interior.
The evidence needs $\lceil 10/4\rceil=3$ chunks and leaves $F=2$ free slots, so $(n_P,n_Q)=(1,1)$: the first frame of the grasp, the ten evidence frames, and the last frame of the grasp form a $12$-frame sequence, cut into three chunks.
If instead $|V|=12$, then $F=0$: $P$ becomes the standalone chunk $\bar{\mathcal S}_4(P)$ (its first and last frame and two interior frames), the evidence fills three chunks, and $Q$, already four frames long, is kept as is.
In both cases the first and last frame of the grasp appear in its chunks (Rule~3), and no chunk contains frames from both the reach and the grasp (Rule~1).

%% file: sec/7_experiments_details.tex
\section{Experimental Details}
\label{app:experiment-protocol}

\subsection{Benchmarks and training data}
\label{app:literature-context}

\paragraph{Implementation and metrics}
Our model is initialized from pretrained LingBot-VA~\citep{li2026causalworldmodelingrobot}.
Event-based preparation retains 1,693 of the 1,750 source demonstrations on DOMINO and 24,844 of the 27,500 source demonstrations on RoboTwin~2.0.
The training schedules use 10k and 50k optimizer updates, respectively, with a learning rate of $10^{-5}$ and an effective batch size of 32.
We report Success Rate (SR) and Manipulation Score (MS) on DOMINO, and SR under clean and random settings on RoboTwin 2.0. 
MS measures terminal spatial progress toward the target with penalties for workspace violations and clutter collisions.
We implement the execution head as a zero-initialized token-wise linear projection, with an execution-loss weight of $\lambda_{\mathrm{exec}}=0.1$ and an inference threshold of $\tau_{\mathrm{exec}}=0.5$.
All inference, following LingBot-VA, use Euler solver with 3 steps for video tokens (integrating to s = 0.6) and 10 steps for action tokens (integrating to s = 1.0).
Execution cost is measured by the number of action vectors issued to the controller.

\paragraph{Benchmark settings}
\Cref{tab:benchmark-protocol} summarizes the benchmark settings and our training schedules. DOMINO evaluation uses clean Level~1. RoboTwin~2.0 evaluation covers both clean and randomized environments.

\begin{table}[h]
\centering
\caption{
Benchmark settings and training schedules for our model.
}
\label{tab:benchmark-protocol}
\begin{tabular}{
    @{}p{0.29\linewidth}
    p{0.27\linewidth}
    p{0.35\linewidth}@{}
}
\toprule
\textbf{Setting} & \textbf{DOMINO} & \textbf{RoboTwin~2.0} \\
\midrule
Tasks
    & 35
    & 50 \\
Source demonstrations
    & 1,750 clean Level~1
    & 2,500 clean + 25,000 randomized \\
Retained demonstrations
    & 1,693
    & 2,465 clean + 22,379 randomized \\
Optimizer updates
    & 10,000
    & 50,000 \\
Learning rate
    & $10^{-5}$
    & $10^{-5}$ \\
Effective batch size
    & 32
    & 64 \\
Evaluation condition
    & Clean Level~1
    & Clean / randomized \\
Evaluation episodes per task
    & 100
    & 100 \\
\bottomrule
\end{tabular}
\end{table}

\paragraph{Dataset preparation}
All event-based models use the same retained demonstration dataset for each benchmark. DOMINO demonstrations come from the benchmark release of \citet{fang2026domino}: 50 clean Level~1 demonstrations for each of the 35 tasks, 1,750 in total. Simulator replay of the released trajectory fails for 57 of them, which are excluded, while the remaining 1,693 are retained. On RoboTwin~2.0, the 27,500 source demonstrations have two origins. The 25,000 randomized demonstrations (500 per task) and the 50 clean demonstrations of \texttt{put\_bottles\_dustbin} are the same source episodes as the released LingBot-VA post-training data \citep{li2026causalworldmodelingrobot}. The released clean demonstrations of the other 49 tasks do not have source trajectories for simulator replay, so we collected 2,450 clean demonstrations (50 per task) ourselves with the RoboTwin~2.0 data generator \citep{chen2025robotwin}. We use 24,844 demonstrations with successful event annotations (2,465 clean and 22,379 randomized). The remaining 2,656 source demonstrations are excluded from this pool because replay failures. \Cref{app:event-annotation} specifies the event annotation process.

\subsection{Model implementation}
\paragraph{Initialization and optimization}
Our model is initialized from the pretrained LingBot-VA base checkpoint. For DOMINO, we use bf16 training with 8 GPUs, a per-gpu batch size of 1, and 4 gradient-accumulation steps, giving an effective batch size of 32. The learning rate is $10^{-5}$, the optimizer moment coefficients are $(\beta_1,\beta_2)=(0.9,0.95)$, weight decay is 0.1, and the warm-up lasts 10 updates. The DOMINO schedule comprises 10,000 updates and the RoboTwin schedule uses 50,000 updates and 64 effective batch size with the same learning rate. We follow LingBot-VA to use three RGB views: one high-mounted camera and two wrist cameras, with configured image dimensions of $256\times320$ (height $\times$ width). We retain the backbone's 30-channel action representation and quantile-based normalization. 

\paragraph{Execution head}
The execution head operates on the normalized, timestep-modulated final action-token features shared with the action output projection. It applies a single linear layer from $d$ features to one execution logit, with parameters shared across all action slots. For $d=3072$, the head introduces only $d+1=3{,}073$ parameters and no additional Transformer blocks.

\paragraph{Temporal grouping}
During training, the number of latent temporal groups per chunk is randomized over $C\in\{1,2,3,4\}$. During inference, we follow LingBot-VA to use $C=2$ for fair comparison. Each future latent temporal group corresponds to 16 action slots. The first inference request has a nominal capacity of 16 future action slots, and subsequent requests have 32.

\paragraph{Noise augmentation and inference}
Following the training setup described in the main experiment, we apply noise augmentation with probability $p=0.5$ and $s_{\mathrm{aug}}\sim\mathcal{U}[0.5,1.0]$. The video and action SNR shifts are 5.0 and 1.0, respectively. For evaluation, we use Euler solver with 3 steps for video tokens (integrating to s = 0.6) and 10 steps for action tokens (integrating to s = 1.0), following the paper statement of LingBot-VA. The classifier-free guidance scales are 5.0 for video and 1.0 for actions.

\subsection{Baselines training configurations}
\label{app:baseline-sources}

\paragraph{Training configurations}
\Cref{tab:local-wam-training} reports the DOMINO training budgets of the WAM baselines. ImageWAM and Fast-WAM both use a learning rate of $10^{-4}$ with a cosine schedule, weight decay of 0.01, a maximum gradient norm of 1.0, bf16 mixed precision, and random seed 42. ImageWAM uses a per-worker batch size of 12 with four accumulation steps; Fast-WAM uses a per-worker batch size of 16 without gradient accumulation. With eight workers, their effective batch sizes are 384 and 128, respectively.

\paragraph{ImageWAM}
We follow the official ImageWAM setting with the FLUX.2-klein-base-4B~\citep{flux2_2025} backbone. Training uses 16-action windows with endpoint-frame visual supervision. The three camera views are packed into a $288\times256$ input, and action/state vectors are 14-dimensional with z-score normalization. The visual and action loss weights are 0.5 and 1.0,  respectively. Both flow schedulers use a shift of 5.0 during training and inference.

\paragraph{Fast-WAM}
We follow the official Fast-WAM setting with the Wan2.2-TI2V-5B~\citep{wan2025wan} backbone. Training uses 32-action windows with an action--video sampling ratio of $4{:}1$. The three camera views are packed into a $384\times320$ input, and action/state vectors are 14-dimensional with z-score normalization. The video and action flow schedulers use shifts of 5.0 and 1.0, respectively, during both training and inference.

\begin{table}[t]
\centering
\caption{\textbf{DOMINO baseline training budgets.}
The last column reports the estimated total number of non-padding training action target occurrences during training, in millions.}
\label{tab:local-wam-training}
\begin{tabular}{@{}lrrlr@{}}
\toprule
\textbf{Method}
    & \textbf{Updates}
    & \textbf{Global batch}
    & \textbf{Sampling unit}
    & \textbf{Action targets (M)} \\
\midrule
LingBot-VA
    & 10,000 & 32
    & Episode segment
    & $\approx55.21$ \\
Fast-WAM
    & 15,000 & 128
    & 32-action window
    & $\approx56.19$ \\
ImageWAM
    & 10,000 & 384
    & 16-action window
    & $\approx58.90$ \\
\bottomrule
\end{tabular}
\end{table}

%% file: sec/8_additional_results.tex
\section{Additional Results}
This appendix reports real-world experiments, task-level results, cost breakdowns, and qualitative examples.

\subsection{Real World Experiments}
\label{sec:exp-realworld}

We conduct simulation to real-world experiments on a Unitree G1 humanoid robot equipped with Unitree Dex1-1 grippers. 
We consider three complex manipulation tasks requiring bimanual coordination:
(1) \textit{Tennis Ball Insertion}, where the robot holds a tube with one arm, inserts a tennis ball with the other, and places the tube upright on a coaster;
(2) \textit{Bimanual Cup Placement}, where the robot sequentially places blue and green cups on their corresponding coasters; and
(3) \textit{Drawer Manipulation}, where the robot opens a drawer, places a cube inside, and closes the drawer.
The policy uses three RGB camera views and predicts
16-dimensional actions comprising 14 arm joint targets and
two gripper commands.

We collect 45, 44, and 90 teleoperated demonstrations for the three tasks, respectively. 
We fine-tune a separate policy for each task for 4k steps, using a learning rate of $10^{-4}$ and a global batch size of 8.
The LingBot-VA baseline is initialized from the official RoboTwin-posttrained checkpoint. 
Our method uses event-aligned sampling and is initialized from our 50k-step RoboTwin checkpoint, including its execution head. 
We maintain the same 3-step video (integrating to s=0.6) and 10-step action (integrating to s=1.0) and $C=2$ for inference.
The evaluation for each method covers 25 independent real-robot rollouts per task.

\input{table/real_world_exp}

\input{fig/real_world_example}

\subsection{Complete task-level results}
\label{app:per-task-results}

\paragraph{DOMINO task-level performance}
\Cref{tab:domino-per-task} reports SR and MS for all 35 DOMINO tasks under the Level~1 setting used in the main evaluation.
It provides the task-level breakdown of \Cref{tab:combined-benchmarks}, where our method achieves 42.83\% SR and 55.92 MS, compared with 32.57\% and 45.55 for LingBot-VA.
\input{table/domino_per_task}

\paragraph{Zero-shot transfer across dynamic levels}
\Cref{tab:domino-zeroshot-l2l3} provides the individual-task results on the ten-task Level~2/3 subset.
The policies are trained on Level~1 and evaluated without target-level adaptation.
At Level~2, our method improves SR from 16.9\% to 29.4\% and MS from 30.37 to 45.42.
At Level~3, SR increases from 6.8\% to 9.7\% and MS increases from 19.36 to 24.23.
The gains of our method generalize beyond the training level, although the low absolute SR at Level~3 show that task completion under these dynamics remains challenging.
\input{table/domino_level23_per_task}

\subsection{Without robot pretraining}
\label{app:generative-init}

We further evaluate our method and LingBot-VA initialized directly from pretrained Wan2.2-TI2V-5B~\citep{wan2025wan} weights, without using LingBot-VA-Base.
As shown in ~\Cref{tab:generative-init}, our method achieves 37.43\% SR and 52.07 MS, compared with 20.83\% SR and 36.57 MS for LingBot-VA, improving SR by 16.60 percentage points and MS by 15.50 points.
Our method also outperforms Fast-WAM and ImageWAM on both metrics, showing that the gains from event-aligned visual action reasoning persist without any robot pretraining.
\input{table/generative_init}

\subsection{Training-sequence composition}
\label{app:sequence-composition}
 
\Cref{tab:sampling-statistics} characterizes the prepared training sequences for the sampling variants in \Cref{tab:main-ablation}.
An action slot is counted each time a position of the demonstration trajectory is sampled, including repetitions.
Slots are classified as inside or outside according to whether their source positions fall within any evidence phase within the interaction event, and percentages use each row's own total.
\input{table/sampling_statistics}

\subsection{Execution cost breakdown}
\label{app:execution-breakdown}
\Cref{tab:paired-execution-detail} expands the execution comparison in~\cref{tab:execution-efficiency}.
For each task, we use the same evaluation episodes on which both LingBot-VA and our model succeed.
$N_{\rm exec}$ counts complete action commands actually issued to the controller.
$N_{\rm roll}$ counts completed future visual-rollout requests.
The paired set contains 738 episodes across all 35 tasks. For each task, we compute each method's mean number of issued action commands over these episodes and the percentage reduction relative to the baseline using these means.
\input{table/paired_execution_table}

\subsection{Additional Visualizations of Event-Aligned Visual Reasoning}
\label{app:qualitative}
Additional visual prediction and attention visualizations in~\cref{fig:appendix_viz1}, ~\cref{fig:appendix_viz2} and~\cref{fig:appendix_viz3} show how visual prediction and executed behavior evolve through manipulation. For visualization of the visual prediction, we integrate to s=1.0 for video tokens here.
Predicted visual states describe the model's anticipated scene evolution, while real observations show the interaction realized during execution. As shown in the predicted visual states, which are the last frame for each chunk, our method proceed the visual reasoning following the interaction dynamics. For each chunk, its visual imagination focuses on one transition or interaction event, which is benefit for action prediction and task completion.
\input{fig/appendix_viz}

%% file: table/real_world_exp.tex
\begin{table}[t]
\centering
\caption{SR(\%) on three real world tasks.}
\label{tab:real-world-exp}
\begin{tabular}{lcccc}
\toprule
\textbf{Method} & \textbf{Tennis} & \textbf{Cup} & \textbf{Drawer} \\
\midrule
LingBot-VA & 44\% & 24\% & 64\% & \\
Ours & 52\% & 28\% & 56\% & \\
\bottomrule
\end{tabular}
\end{table}

%% file: fig/real_world_example.tex
\begin{figure}[t]
    \centering
    \includegraphics[width=0.9\linewidth]{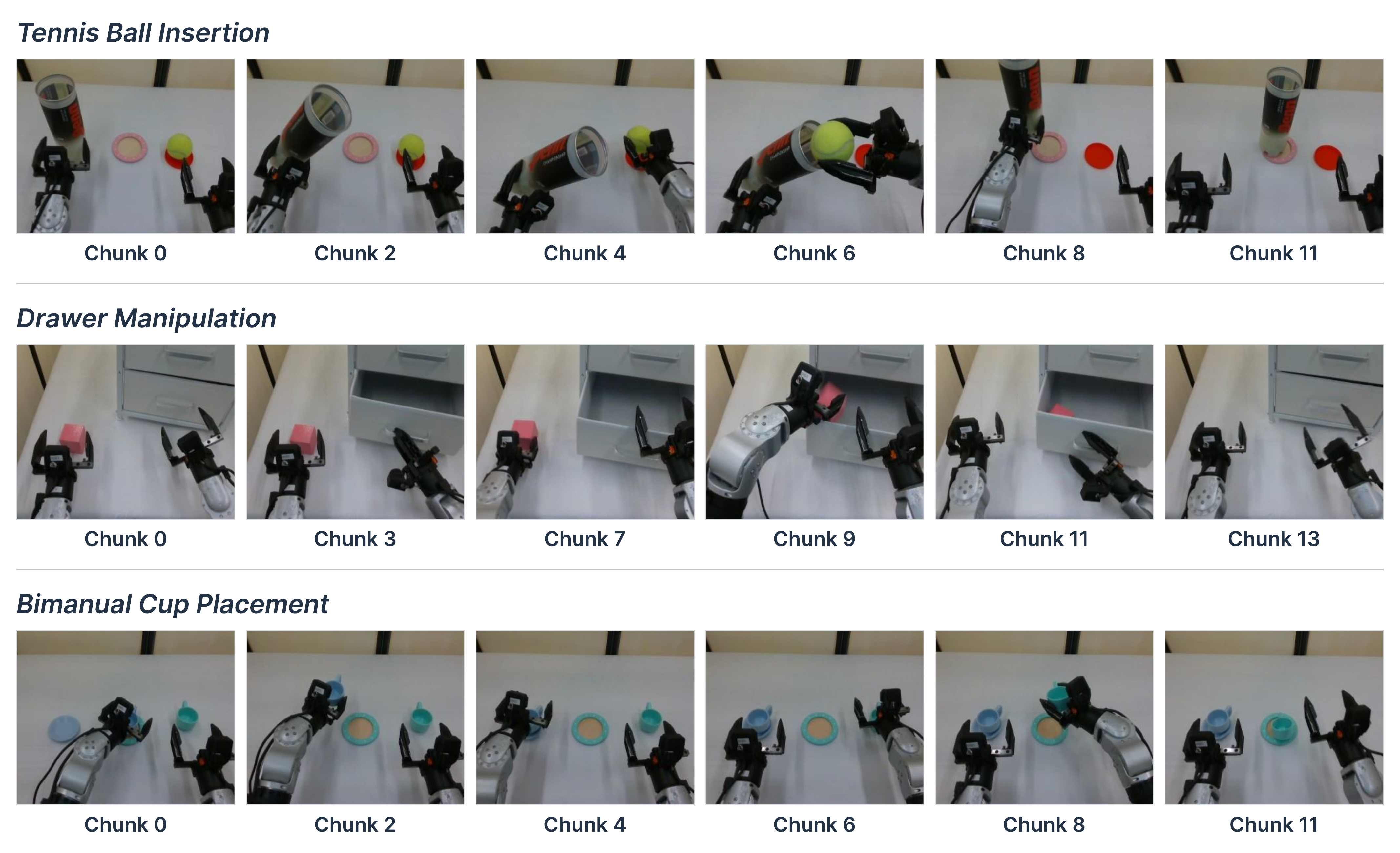}
    \caption{Real world task exmaples.}
    \label{fig:real-world-example}
\end{figure}

%% file: table/domino_per_task.tex
\begin{table}[t]
\centering
\caption{Detailed per-task performance comparison on DOMINO Clean Level~1.}
\begin{tabular}{l*{8}{c}}
\toprule
\multirow{2}{*}{\textbf{Task}} & \multicolumn{2}{c}{\textbf{ImageWAM}}
& \multicolumn{2}{c}{\textbf{Fast-WAM}} & \multicolumn{2}{c}{\textbf{LingBot-VA}}
& \multicolumn{2}{c}{\textbf{Ours}} \\
\cmidrule(lr){2-3}\cmidrule(lr){4-5}\cmidrule(lr){6-7}\cmidrule(lr){8-9}
& \textbf{SR} & \textbf{MS} & \textbf{SR} & \textbf{MS} & \textbf{SR} & \textbf{MS} & \textbf{SR} & \textbf{MS} \\
\midrule
Adjust Bottle & 71 & 80.09 & 63 & 77.53 & \textbf{81} & \textbf{87.51} & 48 & 62.78 \\
Beat Block Hammer & 18 & 31.89 & 12 & 25.45 & 33 & 45.56 & \textbf{49} & \textbf{57.98} \\
Click Alarmclock & 11 & 16.77 & 9 & 13.74 & 9 & 15.24 & \textbf{22} & \textbf{29.51} \\
Click Bell & 2 & 11.65 & 2 & 10.62 & 3 & 14.20 & \textbf{7} & \textbf{18.03} \\
Dump Bin Bigbin & \textbf{36} & 45.86 & \textbf{36} & 44.27 & 33 & \textbf{47.48} & 32 & 38.67 \\
Grab Roller & 48 & 62.63 & 49 & 64.77 & 59 & 68.55 & \textbf{65} & \textbf{77.02} \\
Handover Block & 0 & 38.37 & 9 & 51.90 & \textbf{19} & 56.44 & 13 & \textbf{59.65} \\
Handover Mic & 17 & 42.80 & 37 & 52.83 & \textbf{48} & \textbf{60.06} & 38 & 52.45 \\
Hanging Mug & 7 & 37.74 & 14 & 44.07 & \textbf{39} & \textbf{61.24} & 27 & 58.35 \\
Move Can Pot & 31 & 54.89 & 45 & \textbf{68.03} & \textbf{48} & 63.07 & 32 & 56.99 \\
Move Pillbottle Pad & 17 & 26.41 & 10 & 18.79 & 40 & 45.92 & \textbf{72} & \textbf{75.14} \\
Move Playingcard Away & 6 & 27.27 & 11 & 27.08 & \textbf{30} & \textbf{41.55} & 21 & 32.75 \\
Move Stapler Pad & 3 & 15.31 & 1 & 12.44 & 10 & 21.12 & \textbf{20} & \textbf{33.49} \\
Place A2B Left & 5 & 33.23 & 8 & 34.05 & \textbf{16} & \textbf{38.18} & 13 & 34.35 \\
Place A2B Right & 7 & 33.12 & 10 & 37.72 & \textbf{24} & \textbf{46.17} & 21 & 41.24 \\
Place Bread Basket & 22 & 31.35 & 13 & 23.43 & 41 & 47.67 & \textbf{91} & \textbf{91.76} \\
Place Bread Skillet & 15 & 31.32 & 11 & 27.78 & 11 & 27.75 & \textbf{26} & \textbf{47.85} \\
Place Can Basket & 15 & 40.64 & 17 & 47.06 & \textbf{37} & \textbf{52.66} & 24 & 46.63 \\
Place Container Plate & 37 & 42.19 & 34 & 38.62 & 60 & 63.31 & \textbf{91} & \textbf{91.56} \\
Place Empty Cup & 9 & 19.37 & 5 & 15.03 & 37 & 42.65 & \textbf{62} & \textbf{65.51} \\
Place Fan & 6 & 13.30 & 0 & 9.24 & 14 & 19.88 & \textbf{50} & \textbf{54.90} \\
Place Mouse Pad & 7 & 21.35 & 2 & 17.11 & 19 & 34.00 & \textbf{64} & \textbf{70.19} \\
Place Object Basket & 15 & 39.62 & 18 & 39.53 & 40 & 51.36 & \textbf{72} & \textbf{77.49} \\
Place Object Scale & 7 & 19.57 & 2 & 14.26 & 21 & 30.90 & \textbf{28} & \textbf{39.24} \\
Place Object Stand & 2 & 11.93 & 3 & 13.37 & 17 & 23.45 & \textbf{48} & \textbf{52.42} \\
Place Phone Stand & 9 & 23.94 & 8 & 24.62 & 14 & 29.03 & \textbf{31} & \textbf{41.45} \\
Place Shoe & 26 & 35.40 & 27 & 36.95 & 27 & 39.71 & \textbf{54} & \textbf{63.36} \\
Press Stapler & 9 & 18.41 & 8 & 15.62 & 13 & 22.19 & \textbf{47} & \textbf{52.56} \\
Put Bottles Dustbin & 22 & 36.46 & 35 & 45.53 & \textbf{47} & \textbf{56.99} & 35 & 45.35 \\
Put Object Cabinet & 29 & 59.05 & \textbf{36} & 63.87 & 26 & 53.05 & \textbf{36} & \textbf{64.03} \\
Rotate QRcode & 20 & 37.49 & 15 & 32.17 & 26 & 36.60 & \textbf{34} & \textbf{44.34} \\
Scan Object & 3 & 29.36 & 4 & 28.99 & 11 & 34.77 & \textbf{22} & \textbf{46.08} \\
Shake Bottle & 52 & 67.45 & 53 & 67.06 & 72 & 79.56 & \textbf{73} & \textbf{80.79} \\
Shake Bottle Horizontally & 63 & 78.28 & 55 & 71.69 & \textbf{71} & 82.85 & 68 & \textbf{83.44} \\
Stamp Seal & 13 & 26.93 & 6 & 21.49 & 44 & 53.51 & \textbf{63} & \textbf{70.00} \\
\midrule
\textit{Average} & 18.86 & 35.47 & 19.09 & 35.33 & 32.57 & 45.55 & \textbf{42.83} & \textbf{55.92} \\
\bottomrule
\end{tabular}
\label{tab:domino-per-task}
\end{table}

%% file: table/domino_level23_per_task.tex
\begin{table}[t]
\centering
\caption{Zero-shot transfer to unseen DOMINO dynamic levels.}
\label{tab:domino-zeroshot-l2l3}
\begin{tabular}{l*{8}{c}}
\toprule
\multirow{2}{*}{\textbf{Task}} & \multicolumn{2}{c}{\textbf{ImageWAM}}
& \multicolumn{2}{c}{\textbf{Fast-WAM}} & \multicolumn{2}{c}{\textbf{LingBot-VA}}
& \multicolumn{2}{c}{\textbf{Ours}} \\
\cmidrule(lr){2-3}\cmidrule(lr){4-5}\cmidrule(lr){6-7}\cmidrule(lr){8-9}
& \textbf{SR} & \textbf{MS} & \textbf{SR} & \textbf{MS} & \textbf{SR} & \textbf{MS} & \textbf{SR} & \textbf{MS} \\
\midrule
\multicolumn{9}{c}{\textit{Level 2}} \\
\cmidrule(lr){1-9}
Beat Block Hammer & 2 & 13.52 & 1 & 13.29 & 2 & 13.40 & \textbf{8} & \textbf{20.13} \\
Grab Roller & 40 & 56.81 & 43 & 59.28 & 37 & 50.30 & \textbf{47} & \textbf{65.30} \\
Handover Block & 0 & 36.83 & 8 & 50.49 & \textbf{30} & \textbf{54.91} & 3 & 51.36 \\
Move Pillbottle Pad & 10 & 19.01 & 4 & 13.68 & 10 & 18.52 & \textbf{14} & \textbf{22.34} \\
Place A2B Left & 1 & 30.47 & 1 & 27.48 & \textbf{14} & 32.53 & \textbf{14} & \textbf{32.66} \\
Place Bread Basket & 4 & 16.37 & 3 & 15.87 & 5 & 15.68 & \textbf{41} & \textbf{47.11} \\
Place Can Basket & 13 & 36.96 & \textbf{14} & \textbf{43.48} & 8 & 26.91 & 6 & 32.25 \\
Place Container Plate & 16 & 22.81 & 14 & 20.32 & 13 & 19.60 & \textbf{79} & \textbf{81.03} \\
Press Stapler & 1 & 12.76 & 1 & 12.48 & 3 & 14.18 & \textbf{25} & \textbf{34.06} \\
Shake Bottle & 46 & 58.90 & 46 & 59.23 & 47 & 57.67 & \textbf{57} & \textbf{67.98} \\
\cmidrule(lr){1-9}
\textit{Average} & 13.3 & 30.44 & 13.5 & 31.56 & 16.9 & 30.37 & \textbf{29.4} & \textbf{45.42} \\
\midrule
\multicolumn{9}{c}{\textit{Level 3}} \\
\cmidrule(lr){1-9}
Beat Block Hammer & 2 & 8.03 & 1 & 7.78 & 2 & 10.20 & \textbf{3} & \textbf{12.81} \\
Grab Roller & 9 & 23.70 & 7 & 23.14 & 8 & 23.50 & \textbf{18} & \textbf{33.82} \\
Handover Block & 0 & 28.38 & 4 & 31.09 & \textbf{9} & 36.64 & 4 & \textbf{41.16} \\
Move Pillbottle Pad & 0 & 5.57 & 0 & 5.88 & 4 & 10.47 & \textbf{9} & \textbf{17.75} \\
Place A2B Left & 3 & 17.00 & 2 & 16.65 & \textbf{5} & 18.23 & 4 & \textbf{20.82} \\
Place Bread Basket & 2 & 17.59 & 0 & 16.51 & 6 & 21.64 & \textbf{12} & \textbf{26.62} \\
Place Can Basket & 1 & 16.34 & 2 & 19.43 & 2 & 17.65 & \textbf{7} & \textbf{25.07} \\
Place Container Plate & 4 & 9.23 & 3 & 9.00 & 11 & 16.61 & \textbf{17} & \textbf{23.84} \\
Press Stapler & 0 & 8.41 & 0 & 8.41 & \textbf{3} & 11.57 & \textbf{3} & \textbf{12.81} \\
Shake Bottle & 7 & 16.63 & 5 & 15.52 & 18 & 27.07 & \textbf{20} & \textbf{27.55} \\
\cmidrule(lr){1-9}
\textit{Average} & 2.8 & 15.09 & 2.4 & 15.34 & 6.8 & 19.36 & \textbf{9.7} & \textbf{24.23} \\
\bottomrule
\end{tabular}
\end{table}

%% file: table/generative_init.tex
\begin{table}[t]
    \centering
    \caption{\textbf{Comparison on DOMINO without robot pretraining.}
    Ours and LingBot-VA are initialized directly from Wan2.2-TI2V-5B,
    without using LingBot-VA-Base.}
    \label{tab:generative-init}
    \setlength{\tabcolsep}{8pt}
    \begin{tabular}{@{}llcc@{}}
        \toprule
        Method & Initialization & SR (\%) $\uparrow$ & MS $\uparrow$ \\
        \midrule
        ImageWAM & FLUX.2-klein-base-4B & 18.86 & 35.47 \\
        Fast-WAM & Wan2.2-TI2V-5B & 19.09 & 35.33 \\
        LingBot-VA & Wan2.2-TI2V-5B & 20.83 & 36.57 \\
        \midrule
        \textbf{Ours} & Wan2.2-TI2V-5B & \textbf{37.43} & \textbf{52.07} \\
        \bottomrule
    \end{tabular}
\end{table}

%% file: table/sampling_statistics.tex
\begin{table}[t]
\centering
\caption{Slot composition of sampling configurations in~\cref{tab:main-ablation}.} 
\label{tab:sampling-statistics}
\setlength{\tabcolsep}{4pt}
\begin{tabular}{@{}llrrr@{}}
\toprule
\multirow{1}{*}{\textbf{Sampling}} & \multirow{1}{*}{\textbf{Granularity}}
    & \multirow{1}{*}{\textbf{Slots}}
    & \textbf{Inside evidence phases} & \textbf{Outside evidence phases} \\
\midrule
LingBot-VA (reference)
    & Episode & 291,296 & 94,233 (32.35\%) & 197,063 (67.65\%) \\
\midrule
Episode Uniform
    & Episode & 211,664 & 80,953 (38.25\%) & 130,711 (61.75\%) \\
Event Uniform
    & Event & 211,664 & 100,653 (47.55\%) & 111,011 (52.45\%) \\
\midrule
Reduced Evidence
    & Event & 268,592 & 61,913 (23.05\%) & 206,679 (76.95\%) \\
Reduced Context
    & Event & 173,216 & 116,114 (67.03\%) & 57,102 (32.97\%) \\
\textbf{Event-aligned (Ours)}
    & Event & 211,664 & 116,114 (54.86\%) & 95,550 (45.14\%) \\
\bottomrule
\end{tabular}
\end{table}

%% file: table/paired_execution_table.tex
\begin{table}[t]
\centering
\caption{Execution and visual-rollout requests on common successes.}
\label{tab:paired-execution-detail}
\begin{tabular}{lrrrrrr}
\toprule
\textbf{Task} & \textbf{Paired} \boldmath $n$ & \boldmath $N_{\rm exec}^{\rm base}$ & \boldmath  $N_{\rm exec}^{\rm ours}$ & \textbf{Reduction} (\%) & \boldmath  $N_{\rm roll}^{\rm base}$ & \boldmath  $N_{\rm roll}^{\rm ours}$ \\
\midrule
adjust\_bottle & 42 & 114.40 & 86.02 & 24.81 & 4.83 & 3.50 \\
beat\_block\_hammer & 27 & 111.44 & 77.26 & 30.67 & 4.19 & 3.26 \\
click\_alarmclock & 6 & 68.33 & 75.17 & -10.00 & 3.00 & 3.17 \\
click\_bell & 1 & 216.00 & 77.00 & 64.35 & 8.00 & 3.00 \\
dump\_bin\_bigbin & 13 & 196.85 & 123.38 & 37.32 & 7.15 & 4.85 \\
grab\_roller & 45 & 110.64 & 62.60 & 43.42 & 4.69 & 3.09 \\
handover\_block & 8 & 276.50 & 138.00 & 50.09 & 9.25 & 5.00 \\
handover\_mic & 25 & 243.84 & 119.56 & 50.97 & 8.56 & 4.56 \\
hanging\_mug & 12 & 312.92 & 205.67 & 34.27 & 11.00 & 7.92 \\
move\_can\_pot & 13 & 157.46 & 90.15 & 42.75 & 5.92 & 4.00 \\
move\_pillbottle\_pad & 37 & 139.43 & 92.11 & 33.94 & 5.22 & 4.00 \\
move\_playingcard\_away & 13 & 164.46 & 91.54 & 44.34 & 5.92 & 4.00 \\
move\_stapler\_pad & 4 & 143.50 & 94.75 & 33.97 & 5.50 & 4.25 \\
place\_a2b\_left & 5 & 171.40 & 89.80 & 47.61 & 6.20 & 4.00 \\
place\_a2b\_right & 11 & 156.45 & 102.18 & 34.69 & 5.73 & 4.45 \\
place\_bread\_basket & 35 & 128.54 & 85.97 & 33.12 & 5.00 & 4.00 \\
place\_bread\_skillet & 5 & 126.20 & 86.20 & 31.70 & 5.00 & 4.00 \\
place\_can\_basket & 7 & 361.57 & 216.00 & 40.26 & 12.57 & 8.00 \\
place\_container\_plate & 57 & 135.65 & 88.42 & 34.82 & 5.14 & 4.07 \\
place\_empty\_cup & 33 & 149.21 & 87.52 & 41.35 & 5.85 & 4.09 \\
place\_fan & 9 & 154.89 & 93.89 & 39.38 & 6.00 & 4.00 \\
place\_mouse\_pad & 15 & 165.07 & 129.60 & 21.49 & 6.07 & 5.40 \\
place\_object\_basket & 30 & 282.87 & 182.53 & 35.47 & 9.57 & 7.50 \\
place\_object\_scale & 14 & 136.36 & 97.64 & 28.39 & 5.00 & 4.36 \\
place\_object\_stand & 16 & 130.88 & 87.69 & 33.00 & 5.00 & 4.00 \\
place\_phone\_stand & 12 & 119.25 & 91.92 & 22.92 & 5.00 & 4.00 \\
place\_shoe & 16 & 140.81 & 93.50 & 33.60 & 5.19 & 4.00 \\
press\_stapler & 11 & 92.73 & 61.55 & 33.63 & 3.91 & 2.64 \\
put\_bottles\_dustbin & 24 & 170.21 & 111.58 & 34.44 & 6.25 & 4.54 \\
put\_object\_cabinet & 14 & 256.36 & 254.79 & 0.61 & 9.00 & 9.86 \\
rotate\_qrcode & 15 & 182.60 & 90.27 & 50.57 & 6.80 & 4.00 \\
scan\_object & 6 & 201.00 & 117.17 & 41.71 & 7.33 & 5.17 \\
shake\_bottle & 62 & 90.05 & 67.34 & 25.22 & 3.97 & 3.16 \\
shake\_bottle\_horizontally & 56 & 92.00 & 67.96 & 26.13 & 4.02 & 3.29 \\
stamp\_seal & 39 & 151.59 & 89.36 & 41.05 & 5.92 & 4.18 \\
\midrule
Task mean & 738 & 167.18 & 106.46 & 34.63 & 6.22 & 4.49 \\
\bottomrule
\end{tabular}
\end{table}

%% file: fig/appendix_viz.tex
\begin{figure}
    \centering
    \includegraphics[width=0.9\linewidth]{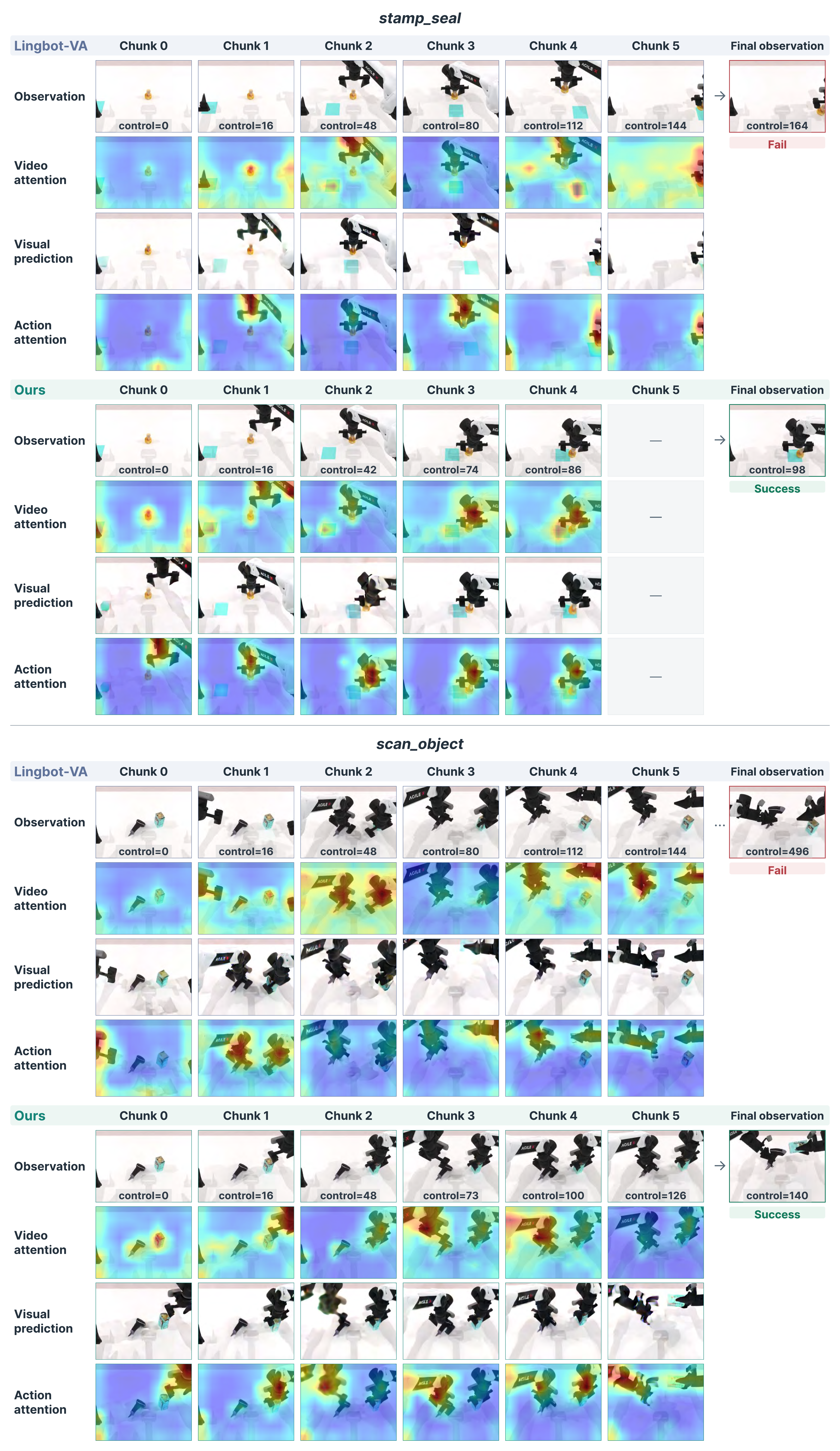}
    \caption{Visual prediction and attention during task execution on \textit{stamp seal} and \textit{scan object}.}
    \label{fig:appendix_viz1}
\end{figure}

\begin{figure}
    \centering
    \includegraphics[width=0.9\linewidth]{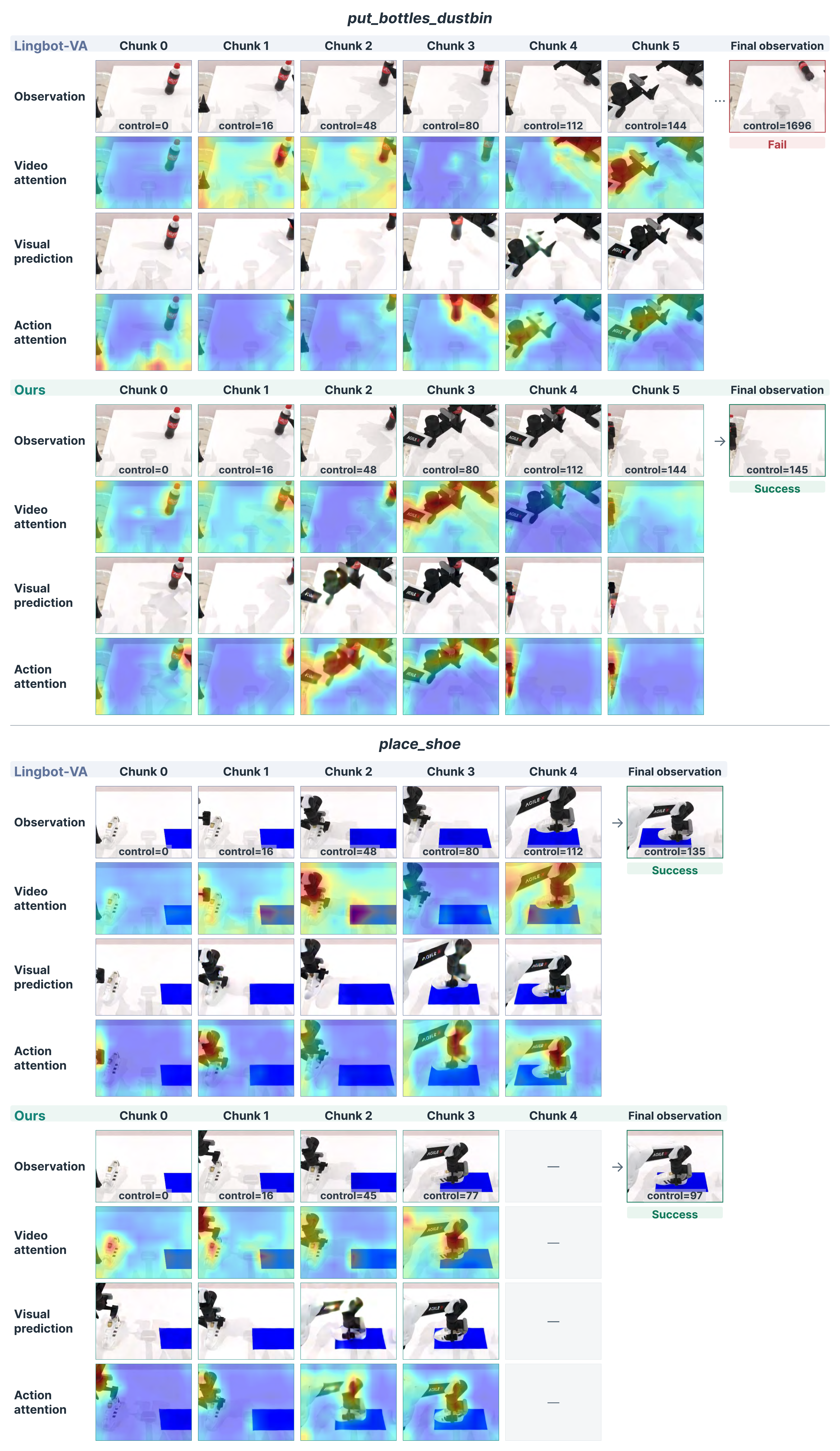}
    \caption{Visual prediction and attention during task execution on \textit{put bottle dustbin} and \textit{place shoe}.}
    \label{fig:appendix_viz2}
\end{figure}

\begin{figure}
    \centering
    \includegraphics[width=0.9\linewidth]{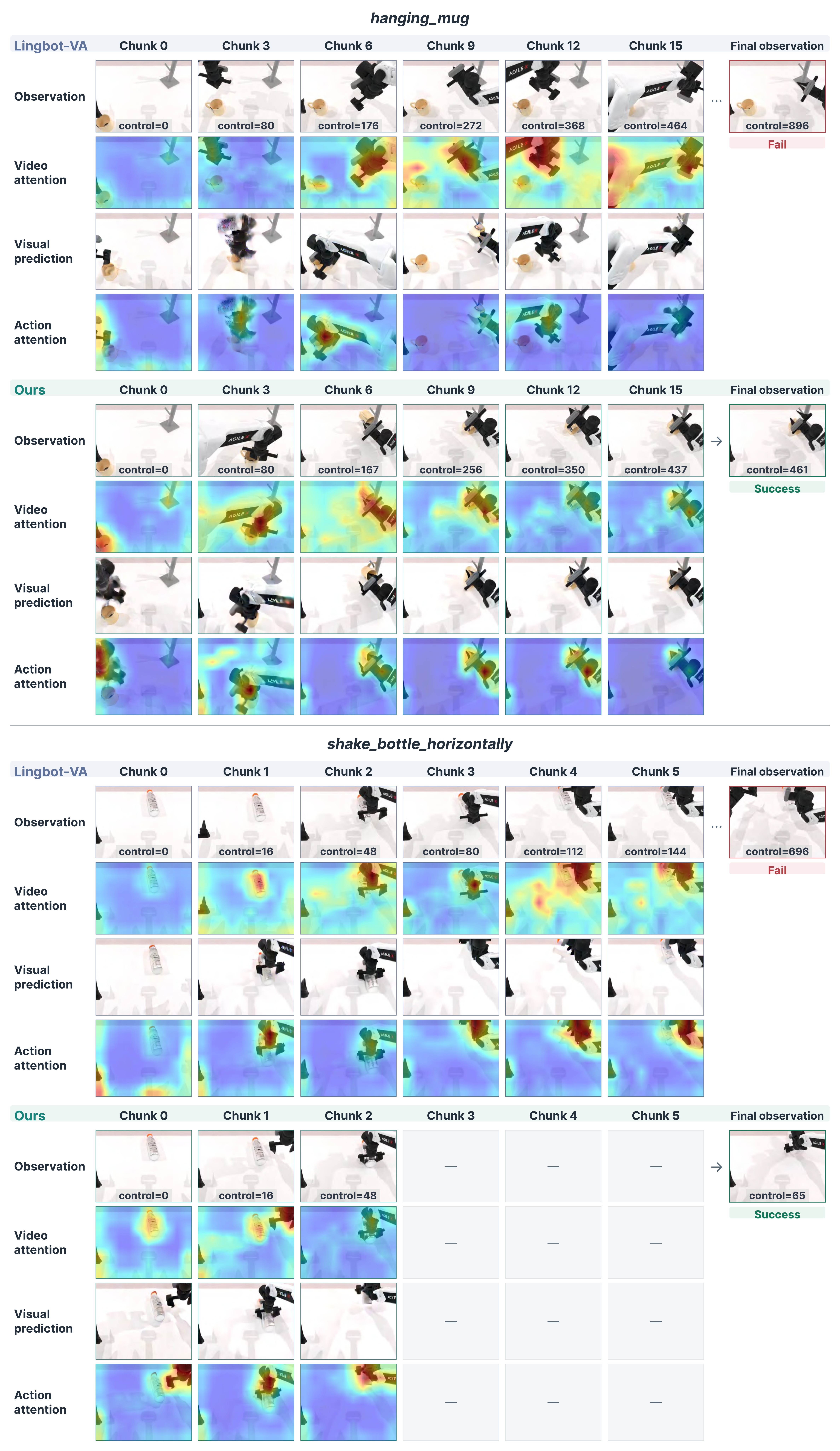}
    \caption{Visual prediction and attention during task execution on \textit{hanging mug} and \textit{shake bottle horizontally}.}
    \label{fig:appendix_viz3}
\end{figure}

%% file: sec/9_event_anotation.tex
\clearpage
\section{Event Annotation for Event-Aligned Visual Reasoning}
\label{app:event-annotation}

This appendix describes the offline annotation pipeline for the event-aligned visual action reasoning framework in \Cref{sec:event_aligned}.
The pipeline first applies reviewed task specifications to each demonstration to localize task-relevant interactions and record their temporal evidence.
The resulting event and phase annotations then provide the inputs to event-aligned visual-action chunk construction.
We describe this process in the order of task specifications, episode-level temporal annotation, and the interface to chunk construction.

\subsection{Task specifications and annotation inputs}
\label{app:event-provenance}

\paragraph{Task specifications}
We used an LLM (Claude-Opus 5) to draft task-level event specifications from simulator source code, including scene construction, expert execution procedures, and success predicates. 
Each draft was manually inspected and corrected before use. 
The reviewed specifications cover 35 DOMINO and 50 RoboTwin~2.0 tasks and identify the relevant objects and arms, native-predicate mappings, and applicable detection rules, with functional contact pairs specified where applicable.
Each specification is reused across demonstrations of the same task.
Episode-specific event timestamps were subsequently extracted during simulator replay by applying event detectors configured using the reviewed specifications. 
The prompt template used to draft the specifications is provided in \Cref{app:task-specification-prompt}.

\paragraph{Annotation inputs}
The simulator supplies physical states and contact reports, and the benchmark task implementations supply native success predicates and expert routines.
Our annotation protocol defines the derived contact and hold conditions, detector labels, and temporal confirmation rules applied to these inputs.
Native success predicates retain the conditions specified by each benchmark.
\Cref{tab:event-provenance} summarizes the provenance and use of these inputs.

\begin{table}[t]
\centering
\small
\caption{\textbf{Inputs to the event annotation pipeline.}
Reviewed task specifications select the entities and detection rules.
Measurements and recorded execution determine the times for each demonstration.}
\label{tab:event-provenance}
\begin{tabularx}{\linewidth}{@{}>{\raggedright\arraybackslash}p{0.22\linewidth}>{\raggedright\arraybackslash}p{0.28\linewidth}>{\raggedright\arraybackslash}X@{}}
\toprule
\textbf{Input} & \textbf{Provenance} & \textbf{Use in annotation} \\
\midrule
Physical states and contacts &
Recorded robot values and replayed simulator state &
Identify gripper changes, object contact and motion, relative-pose stability, and articulation changes. \\
Task-success predicates &
Native \texttt{check\_\allowbreak{}success()} in benchmark task code &
Localize task-specific state changes using the original predicate and our temporal confirmation rule. \\
Expert execution intervals &
Recorded execution of benchmark expert routines &
Localize designated task motions, such as the DOMINO shaking routines. \\
Task specifications and detection rules &
Our annotation protocol, with LLM-assisted specifications reviewed task by task &
Select actors, contact pairs, detector mappings, and the rules used to localize and verify interactions. \\
\bottomrule
\end{tabularx}
\end{table}

\subsection{Event localization and temporal annotation}
\label{app:event-timeline}

\paragraph{Source timeline}
Let $L_{\mathrm{src}}$ denote the number of observations in a raw demonstration, indexed by $t\in\{0,\ldots,L_{\mathrm{src}}-1\}$.
Replay records robot and scene states at the original observation-capture boundaries and evaluates the native task-success predicate on the same timeline.
All annotation times and temporal windows refer to these source observations, before resampling or VAE encoding.

\paragraph{Localizing interactions}
The reviewed specification selects the detectors applied to each demonstration.
Their inputs and outputs are summarized below:
\begin{itemize}
    \item \emph{Acquisition and release.}
    Shared rules use recorded gripper values, replayed contacts, relative poses, and object motion to identify object grasps and releases.
    Handover and bimanual records combine the corresponding arm-specific detections.
    \item \emph{Functional contact.}
    Contacts between configured body pairs localize interactions such as placement, hanging, and stacking.
    Native task predicates can also be mapped to this detector category. Each record retains its source.
    \item \emph{Task-specific state changes.}
    Native task predicates identify conditions such as reaching a target region or satisfying an orientation requirement.
    Articulated-object detections additionally use contact, pose, and joint-state measurements.
    \item \emph{Recorded task motions.}
    Designated expert-execution intervals provide temporal anchors for motions such as DOMINO's vertical and horizontal shaking routines.
\end{itemize}
The configured detector subtypes for each task are listed in \Cref{app:event-inventory}.

\paragraph{Event times}
Each detector record identifies the detected change, its participating objects and arms, and three temporal anchors: onset $t_{\mathrm{on}}$, effect $t_{\mathrm{eff}}$, and confirmation $t_{\mathrm{conf}}$, with
\begin{equation}
    0\leq t_{\mathrm{on}}\leq t_{\mathrm{eff}}\leq
    t_{\mathrm{conf}}<L_{\mathrm{src}}.
    \label{eq:event-record}
\end{equation}
Onset marks the beginning of the qualifying condition, effect anchors the detector-specific change, and confirmation records its verification anchor.
For ordinary and handle grasps, the saved confirmation time marks the start of a qualifying stability window. 
For releases, handover receives, contacts, and success predicates, it marks the final confirmation row.
The record therefore also retains the complete verification window and the measurements used by the detector.

\paragraph{Detector evidence windows}
\label{app:evidence-windows}
For a detector record $\eta$, we recover a half-open window $[u_\eta,v_\eta)$ containing the source observations used to verify the detected interaction.
These windows follow the saved contact, stability, motion, or expert-execution records, rather than a fixed-radius neighborhood around the effect time.
In particular, grasp verification can extend beyond the saved confirmation anchor.
\Cref{tab:evidence-windows} summarizes the window sources in the DOMINO annotation protocol.

\begin{table}[t]
\centering
\small
\caption{\textbf{Detector evidence windows in DOMINO.}
Windows retain the source observations used for verification.
Their relation to the method's event and phase annotations is described in \Cref{app:event-boundaries}.}
\label{tab:evidence-windows}
\begin{tabularx}{\linewidth}{@{}>{\raggedright\arraybackslash}p{0.29\linewidth}>{\raggedright\arraybackslash}X@{}}
\toprule
\textbf{Detector source} & \textbf{Temporal evidence retained} \\
\midrule
Stable closed contact &
From onset through the saved post-evidence end, including the complete qualifying stability window. \\
Merged bimanual grasp &
From onset through the latest of the merged confirmation row and both component post-evidence windows. \\
Handover receive; object release / handover give &
$[t_{\mathrm{on}},t_{\mathrm{conf}}+1)$, including the detector's stability or release-confirmation observations. \\
Task-specific body-pair contact &
From contact onset through confirmation and the associated task-success or object-release effect, as specified by the detector. \\
Handle hold and drawer motion &
The linked drawer record supplies the held-run window for handle verification; drawer-motion evidence spans the saved baseline-to-peak-search interval. \\
Native success or expert motion &
$[t_{\mathrm{on}},t_{\mathrm{conf}}+1)$, using the predicate-confirmation interval or the designated expert-method record. \\
\bottomrule
\end{tabularx}
\end{table}

\subsection{From temporal annotations to chunk construction}
\label{app:event-boundaries}

\paragraph{Event and phase annotations}
Detector times provide temporal anchors for the event annotations in \Cref{sec:event_aligned}.
Gripper opening and closing starts provide additional anchors for the surrounding manipulation motion.
Records can overlap or describe different aspects of the same interaction, so detector records are distinct from the contiguous event partition used by the method.
In particular, an effect time alone does not specify both boundaries of an interaction event.

The annotation interface to chunk construction consists of the events $(E_i,c_i)$ with boundaries $(s_i,e_i)$, together with evidence boundaries $(s_i^\mathrm{ev},e_i^\mathrm{ev})$ for each interaction event.
The detector windows retain the measurements used to verify a change, while the evidence and support phases specify the temporal organization of visual states within the interaction.
The chunk constructor receives these event and phase boundaries as fixed annotations on the source timeline, following the interaction taxonomy in \Cref{sec:event_aligned}.

\paragraph{Retained trajectory}
In the DOMINO sampling protocol, the retained trajectory includes the last event anchor and any later observations required by the saved detector evidence.
Extending this final sampling range preserves the original annotation times.
Row 0 remains the initial observation, and unused trailing observations are omitted.
The method and its matched uniform sampling controls use the same retained endpoint.
The final retained frame is $T$ in the notation of \Cref{sec:event_aligned}; $L_{\mathrm{src}}$ denotes the raw observation count.

\paragraph{Visual-action targets}
The annotated events and evidence boundaries are passed to \Cref{alg:event-aligned-chunks} to construct the visual-action training targets.
Event boundaries determine chunk membership, and the evidence and support phases determine the temporal sampling within an interaction.
Actions remain paired with their selected visual frames, and repeated positions provide supervision for the execution validity head in \Cref{sec:validity_head}.
The construction operators and cases are detailed in \Cref{app:construction-detail}.

\subsection{Annotation checks and dataset selection}
\label{app:event-quality}

\paragraph{Consistency checks}
We check replay alignment with the source demonstration and verify that detector labels and temporal records are consistent with the reviewed task specification and recorded execution.
Replay failures and insufficient evidence are recorded explicitly.
The annotation manifest retains episode status and any replay discrepancies used in downstream data selection.

\paragraph{DOMINO data selection}
The audited DOMINO manifest contains $35\times50=1{,}750$ source demonstrations, of which 1,693 are retained.
These include 1,364 with successful canonical annotations, 328 with recorded end-effector replay discrepancies, and one with an explicitly recorded short-confirmation exception using two terminal true observations instead of the default three.
The remaining 57 demonstrations have replay failures and are excluded.

\paragraph{Use of annotation inputs}
Simulator states, contacts, native task predicates, and expert-method markers are used offline to construct supervision.
Their role is to localize interactions in the source demonstrations.
Inference uses the learned visual-action model and execution validity predictions described in the main text.

\subsection{Prompt for task-level event specifications}
\label{app:task-specification-prompt}

We used the following prompt with Claude Opus 5 to draft the task-level event specifications.
\begin{tcblisting}{
  listing engine=listings,
  listing only,    
  breakable,
  colback=gray!4,
  colframe=gray!70!black,
  boxrule=0.5pt,
  arc=2pt,
  left=4pt, right=4pt, top=3pt, bottom=3pt,
  title={Prompt for drafting task-level event specifications},
  fonttitle=\small\bfseries,
  listing options={
    language={},
    basicstyle=\ttfamily\footnotesize\color{black},
    keywordstyle={},
    identifierstyle={},
    commentstyle={},
    stringstyle={},
    showstringspaces=false,
    showspaces=false,
    showtabs=false,
    numbers=none,
    frame=none,
    columns=fullflexible,
    keepspaces=true,
    breaklines=true,
    breakatwhitespace=true,
    breakautoindent=false,
    breakindent=1em,
    tabsize=2,
    xleftmargin=0pt,
    xrightmargin=0pt,
    aboveskip=0pt,
    belowskip=0pt,
  }
}
You are assisting in constructing task-level event specifications for robotic manipulation datasets. Your task is to draft a structured specification from the supplied simulator source code.

INPUTS
- Platform: {DOMINO or RoboTwin 2.0}
- Task source files: {envs/<task_name>.py}
- Supported specification schema and event-detector conventions:
  {schema_and_detector_contract}

SOURCE ANALYSIS
For each task, inspect:
1. Scene construction, such as load_actors(), to identify actor attributes and their meanings.
2. The expert execution procedure, such as play_once(), including conditional branches, handovers, regrasps, and bimanual actions.
3. check_success() and any helper functions it calls to determine the actual success conditions.

SPECIFICATION RULES
1. Use exact environment attribute names for:
   - grasp_actor_attributes: candidate actors for grasp detection;
   - primary_actor_attributes: actors associated with the main event.
2. Set expected_grasp_min using the expert procedure and the supplied detector conventions.
3. Define primary_event using the supported event vocabulary:
   - type: event category;
   - subtype: a concise description of the transition or interaction;
   - role: its semantic role in task execution;
   - predicate_definition: a faithful English summary of check_success().
4. Preserve the predicate's actual logic, including thresholds, coordinate frames, logical operators, branch conditions, and persistent or latched state. Do not add contact, release, stability, orientation, or displacement requirements unless the source checks them.
5. Add optional fields, such as functional_pairs, grasp_mode, handover, or branch-specific requirements, only when supported by both the source and the supplied detector contract.
6. Do not invent event timestamps or infer rules from task names alone.

OUTPUT EXAMPLE
{
  "source": {
    "platform": "RoboTwin 2.0",
    "task_module_root": "envs"
  },
  "event_contract": {
    "index_space": "episode-local source observation row",
    "model_independent": true
  },
  "tasks": {
    "adjust_bottle": {
      "source_file": "envs/adjust_bottle.py",
      "grasp_actor_attributes": ["bottle"],
      "primary_actor_attributes": ["bottle"],
      "expected_grasp_min": 1,
      "primary_event": {
        "type": "state_transition",
        "subtype": "high_at_instructed_side",
        "role": "COMPLETE",
        "predicate_definition": "Bottle functional point 0 is above world Z 0.9 and is left of world X -0.15 for qpose_tag 0 or right of world X 0.15 for qpose_tag 1; gripper state, contact, and displacement are not checked."
      }
    }
  }
}
\end{tcblisting}

\subsection{Task-specific detector inventory}
\input{table/event_task_inventory}

\label{app:event-inventory}

\Cref{tab:event-domino-1} lists the DOMINO task mappings, and \Cref{tab:event-robotwin-1,tab:event-robotwin-2} list the RoboTwin~2.0 mappings.
Prefixes S, F, and G denote detector categories: \texttt{state\_\allowbreak{}transition}, \texttt{functional\_\allowbreak{}contact}, and \texttt{grasp}, respectively.
These categories identify detected changes. The method uses the interaction and transition event types defined in \Cref{sec:event_aligned}.
In particular, the detector label \texttt{state\_\allowbreak{}transition} can mark a task-relevant interaction.
Generic object-grasp, release, and gripper-change labels are shared across tasks and are omitted from individual rows.
Additional G entries specify bimanual, handle, or handover-receive detections, with receives paired with \texttt{release/handover\_release}.
The inventory lists configured subtypes.

%% file: table/event_task_inventory.tex
\begin{table}[t]
\centering
\small
\caption{DOMINO task-to-detector-subtype mapping.
Prefixes S, F, and G identify detector categories.}
\label{tab:event-domino-1}
\begin{tabularx}{\linewidth}{@{}>{\raggedright\arraybackslash}p{0.38\linewidth}>{\raggedright\arraybackslash}X@{}}
\toprule
\textbf{Task(s)} & \textbf{Configured detector subtype(s) and interpretation} \\
\midrule
\texttt{adjust\_\allowbreak{}bottle} & S: \texttt{high\_\allowbreak{}at\_\allowbreak{}side}\par Bottle reaches the required side and height. \\[3pt]
\texttt{beat\_\allowbreak{}block\_\allowbreak{}hammer} & F: \texttt{hit}\par Hammer satisfies the task-specific hit condition. \\[3pt]
\texttt{click\_\allowbreak{}alarmclock}, \texttt{click\_\allowbreak{}bell}, \texttt{press\_\allowbreak{}stapler} & F: \texttt{press}\par Gripper satisfies the task-specific press condition. \\[3pt]
\texttt{dump\_\allowbreak{}bin\_\allowbreak{}bigbin} & S: \texttt{contents\_\allowbreak{}height\_\allowbreak{}band}\par Bin reaches the required height; waste enters the height band. \\[3pt]
\texttt{grab\_\allowbreak{}roller} & S: \texttt{lifted\_\allowbreak{}bimanually};\allowbreak{} G: \texttt{bimanual\_\allowbreak{}grasp}\par Two-arm acquisition and the roller-lift predicate. \\[3pt]
\texttt{handover\_\allowbreak{}block} & S: \texttt{placed\_\allowbreak{}at\_\allowbreak{}target};\allowbreak{} G: \texttt{handover\_\allowbreak{}receive}\par Handover followed by placement at the target. \\[3pt]
\texttt{handover\_\allowbreak{}mic} & S: \texttt{handover\_\allowbreak{}terminal\_\allowbreak{}configuration};\allowbreak{} G: \texttt{handover\_\allowbreak{}receive}\par Handover reaches the required terminal configuration. \\[3pt]
\texttt{hanging\_\allowbreak{}mug} & S: \texttt{hung};\allowbreak{} F: \texttt{hang\_\allowbreak{}contact}\par Mug reaches the rack, is released, and contacts it. \\[3pt]
\texttt{move\_\allowbreak{}can\_\allowbreak{}pot}, \texttt{place\_\allowbreak{}a2b\_\allowbreak{}left}, \texttt{place\_\allowbreak{}a2b\_\allowbreak{}right} & S: \texttt{placed\_\allowbreak{}relative}\par Relative-position and release conditions hold. \\[3pt]
\texttt{move\_\allowbreak{}pillbottle\_\allowbreak{}pad}, \texttt{place\_\allowbreak{}container\_\allowbreak{}plate}, \texttt{place\_\allowbreak{}empty\_\allowbreak{}cup}, \texttt{place\_\allowbreak{}object\_\allowbreak{}scale}, \texttt{place\_\allowbreak{}object\_\allowbreak{}stand}, \texttt{place\_\allowbreak{}phone\_\allowbreak{}stand}, \texttt{stamp\_\allowbreak{}seal} & S: \texttt{placed};\allowbreak{} F: \texttt{place\_\allowbreak{}contact}\par Placement succeeds with contact at the designated target. \\[3pt]
\texttt{move\_\allowbreak{}playingcard\_\allowbreak{}away} & S: \texttt{moved\_\allowbreak{}away}\par Card leaves the central X band and satisfies release and applicable displacement conditions. \\[3pt]
\texttt{move\_\allowbreak{}stapler\_\allowbreak{}pad}, \texttt{place\_\allowbreak{}fan}, \texttt{place\_\allowbreak{}mouse\_\allowbreak{}pad}, \texttt{place\_\allowbreak{}shoe} & S: \texttt{placed\_\allowbreak{}oriented};\allowbreak{} F: \texttt{place\_\allowbreak{}contact}\par Placement satisfies position, orientation, and contact conditions. \\[3pt]
\texttt{place\_\allowbreak{}bread\_\allowbreak{}basket} & S: \texttt{contacting\_\allowbreak{}target\_\allowbreak{}region};\allowbreak{} F: \texttt{place\_\allowbreak{}contact}\par Bread satisfies basket-region, contact, and release conditions. \\[3pt]
\texttt{place\_\allowbreak{}bread\_\allowbreak{}skillet} & S: \texttt{positioned\_\allowbreak{}at\_\allowbreak{}target};\allowbreak{} F: \texttt{place\_\allowbreak{}contact}\par Bread reaches the skillet target region; the primary predicate does not require release. \\[3pt]
\texttt{place\_\allowbreak{}can\_\allowbreak{}basket}, \texttt{place\_\allowbreak{}object\_\allowbreak{}basket} & S: \texttt{contacting\_\allowbreak{}lifted\_\allowbreak{}target};\allowbreak{} F: \texttt{place\_\allowbreak{}contact}\par Object and lifted basket satisfy spatial and contact conditions. \\[3pt]
\texttt{put\_\allowbreak{}bottles\_\allowbreak{}dustbin} & S: \texttt{fixed\_\allowbreak{}region\_\allowbreak{}entry};\allowbreak{} G: \texttt{handover\_\allowbreak{}receive}\par Bottle enters the target region; handover depends on the source branch. \\[3pt]
\texttt{put\_\allowbreak{}object\_\allowbreak{}cabinet} & S: \texttt{target\_\allowbreak{}region\_\allowbreak{}entry};\allowbreak{} S: \texttt{drawer\_\allowbreak{}opened};\allowbreak{} G: \texttt{handle\_\allowbreak{}grasp}\par Drawer motion followed by entry into the cabinet target region. \\[3pt]
\texttt{rotate\_\allowbreak{}qrcode} & S: \texttt{oriented}\par QR-code object meets orientation, height, and applicable position conditions. \\[3pt]
\texttt{scan\_\allowbreak{}object} & S: \texttt{aligned}\par Object satisfies scanner-axis alignment and distance conditions. \\[3pt]
\texttt{shake\_\allowbreak{}bottle} & S: \texttt{lift\_\allowbreak{}proxy};\allowbreak{} S: \texttt{shaken}\par Lift proxy plus the recorded completion of the expert vertical-shake method. \\[3pt]
\texttt{shake\_\allowbreak{}bottle\_\allowbreak{}horizontally} & S: \texttt{lift\_\allowbreak{}proxy};\allowbreak{} S: \texttt{shaken\_\allowbreak{}horizontally}\par Lift proxy plus the recorded completion of the expert horizontal-shake method. \\[3pt]
\bottomrule
\end{tabularx}
\end{table}

\begin{table}[t]
\small
\centering
\caption{RoboTwin~2.0 task-to-detector-subtype mapping (1/2).
Prefixes S, F, and G identify detector categories.}
\small
\label{tab:event-robotwin-1}
\begin{tabularx}{\linewidth}{@{}>{\raggedright\arraybackslash}p{0.38\linewidth}>{\raggedright\arraybackslash}X@{}}
\toprule
\textbf{Task(s)} & \textbf{Configured detector subtype(s)} \\
\midrule
\texttt{adjust\_\allowbreak{}bottle} & S: \texttt{high\_\allowbreak{}at\_\allowbreak{}instructed\_\allowbreak{}side} \\[3pt]
\texttt{beat\_\allowbreak{}block\_\allowbreak{}hammer} & F: \texttt{hit};\allowbreak{} F: \texttt{hit\_\allowbreak{}contact} \\[3pt]
\texttt{blocks\_\allowbreak{}ranking\_\allowbreak{}rgb} & S: \texttt{ordered\_\allowbreak{}red\_\allowbreak{}green\_\allowbreak{}blue} \\[3pt]
\texttt{blocks\_\allowbreak{}ranking\_\allowbreak{}size} & S: \texttt{ordered\_\allowbreak{}large\_\allowbreak{}medium\_\allowbreak{}small} \\[3pt]
\texttt{click\_\allowbreak{}alarmclock}, \texttt{click\_\allowbreak{}bell}, \texttt{press\_\allowbreak{}stapler} & F: \texttt{press} \\[3pt]
\texttt{dump\_\allowbreak{}bin\_\allowbreak{}bigbin} & S: \texttt{contents\_\allowbreak{}height\_\allowbreak{}band} \\[3pt]
\texttt{grab\_\allowbreak{}roller} & S: \texttt{lifted\_\allowbreak{}with\_\allowbreak{}closed\_\allowbreak{}grippers};\allowbreak{} G: \texttt{bimanual\_\allowbreak{}grasp} \\[3pt]
\texttt{handover\_\allowbreak{}block} & S: \texttt{placed\_\allowbreak{}at\_\allowbreak{}target};\allowbreak{} F: \texttt{support\_\allowbreak{}contact};\allowbreak{} G: \texttt{handover\_\allowbreak{}receive} \\[3pt]
\texttt{handover\_\allowbreak{}mic} & S: \texttt{handover\_\allowbreak{}terminal\_\allowbreak{}configuration};\allowbreak{} G: \texttt{handover\_\allowbreak{}receive} \\[3pt]
\texttt{hanging\_\allowbreak{}mug} & S: \texttt{hung\_\allowbreak{}geometry};\allowbreak{} F: \texttt{hang\_\allowbreak{}contact} \\[3pt]
\texttt{lift\_\allowbreak{}pot} & S: \texttt{lifted\_\allowbreak{}upright\_\allowbreak{}between\_\allowbreak{}tcps};\allowbreak{} G: \texttt{bimanual\_\allowbreak{}grasp} \\[3pt]
\texttt{move\_\allowbreak{}can\_\allowbreak{}pot} & S: \texttt{placed\_\allowbreak{}relative} \\[3pt]
\texttt{move\_\allowbreak{}pillbottle\_\allowbreak{}pad} & S: \texttt{placed\_\allowbreak{}on\_\allowbreak{}pad\_\allowbreak{}geometry};\allowbreak{} F: \texttt{support\_\allowbreak{}contact} \\[3pt]
\texttt{move\_\allowbreak{}playingcard\_\allowbreak{}away} & S: \texttt{outside\_\allowbreak{}center\_\allowbreak{}x\_\allowbreak{}band} \\[3pt]
\texttt{move\_\allowbreak{}stapler\_\allowbreak{}pad}, \texttt{place\_\allowbreak{}fan}, \texttt{place\_\allowbreak{}mouse\_\allowbreak{}pad} & S: \texttt{placed\_\allowbreak{}on\_\allowbreak{}pad\_\allowbreak{}oriented};\allowbreak{} F: \texttt{support\_\allowbreak{}contact} \\[3pt]
\texttt{open\_\allowbreak{}laptop} & S: \texttt{joint\_\allowbreak{}threshold\_\allowbreak{}with\_\allowbreak{}tcp\_\allowbreak{}proximity} \\[3pt]
\texttt{open\_\allowbreak{}microwave} & S: \texttt{joint\_\allowbreak{}threshold} \\[3pt]
\texttt{pick\_\allowbreak{}diverse\_\allowbreak{}bottles}, \texttt{pick\_\allowbreak{}dual\_\allowbreak{}bottles} & S: \texttt{dual\_\allowbreak{}bottles\_\allowbreak{}at\_\allowbreak{}lift\_\allowbreak{}targets} \\[3pt]
\texttt{place\_\allowbreak{}a2b\_\allowbreak{}left} & S: \texttt{placed\_\allowbreak{}relative\_\allowbreak{}left} \\[3pt]
\texttt{place\_\allowbreak{}a2b\_\allowbreak{}right} & S: \texttt{placed\_\allowbreak{}relative\_\allowbreak{}right} \\[3pt]
\texttt{place\_\allowbreak{}bread\_\allowbreak{}basket} & S: \texttt{all\_\allowbreak{}bread\_\allowbreak{}in\_\allowbreak{}basket\_\allowbreak{}region};\allowbreak{} F: \texttt{containment\_\allowbreak{}contact} \\[3pt]
\texttt{place\_\allowbreak{}bread\_\allowbreak{}skillet} & S: \texttt{positioned\_\allowbreak{}over\_\allowbreak{}lifted\_\allowbreak{}skillet};\allowbreak{} F: \texttt{support\_\allowbreak{}contact} \\[3pt]
\bottomrule
\end{tabularx}
\end{table}

\begin{table}[t]
\small
\centering
\caption{RoboTwin~2.0 task-to-detector-subtype mapping (2/2).
Prefixes S, F, and G identify detector categories.}
\small
\label{tab:event-robotwin-2}
\begin{tabularx}{\linewidth}{@{}>{\raggedright\arraybackslash}p{0.38\linewidth}>{\raggedright\arraybackslash}X@{}}
\toprule
\textbf{Task(s)} & \textbf{Configured detector subtype(s)} \\
\midrule
\texttt{place\_\allowbreak{}burger\_\allowbreak{}fries} & S: \texttt{both\_\allowbreak{}items\_\allowbreak{}at\_\allowbreak{}tray\_\allowbreak{}regions};\allowbreak{} F: \texttt{support\_\allowbreak{}contact} \\[3pt]
\texttt{place\_\allowbreak{}can\_\allowbreak{}basket}, \texttt{place\_\allowbreak{}object\_\allowbreak{}basket} & S: \texttt{contacting\_\allowbreak{}lifted\_\allowbreak{}basket};\allowbreak{} F: \texttt{containment\_\allowbreak{}contact} \\[3pt]
\texttt{place\_\allowbreak{}cans\_\allowbreak{}plasticbox} & S: \texttt{both\_\allowbreak{}cans\_\allowbreak{}in\_\allowbreak{}box\_\allowbreak{}regions};\allowbreak{} F: \texttt{containment\_\allowbreak{}contact} \\[3pt]
\texttt{place\_\allowbreak{}container\_\allowbreak{}plate} & S: \texttt{placed\_\allowbreak{}on\_\allowbreak{}plate\_\allowbreak{}geometry};\allowbreak{} F: \texttt{support\_\allowbreak{}contact} \\[3pt]
\texttt{place\_\allowbreak{}dual\_\allowbreak{}shoes} & S: \texttt{paired\_\allowbreak{}shoes\_\allowbreak{}in\_\allowbreak{}box};\allowbreak{} F: \texttt{containment\_\allowbreak{}contact} \\[3pt]
\texttt{place\_\allowbreak{}empty\_\allowbreak{}cup} & S: \texttt{placed\_\allowbreak{}on\_\allowbreak{}coaster\_\allowbreak{}geometry};\allowbreak{} F: \texttt{support\_\allowbreak{}contact} \\[3pt]
\texttt{place\_\allowbreak{}object\_\allowbreak{}scale} & S: \texttt{placed\_\allowbreak{}on\_\allowbreak{}scale\_\allowbreak{}geometry};\allowbreak{} F: \texttt{support\_\allowbreak{}contact} \\[3pt]
\texttt{place\_\allowbreak{}object\_\allowbreak{}stand} & S: \texttt{placed\_\allowbreak{}on\_\allowbreak{}stand\_\allowbreak{}geometry};\allowbreak{} F: \texttt{support\_\allowbreak{}contact} \\[3pt]
\texttt{place\_\allowbreak{}phone\_\allowbreak{}stand} & S: \texttt{placed\_\allowbreak{}on\_\allowbreak{}phone\_\allowbreak{}stand\_\allowbreak{}geometry};\allowbreak{} F: \texttt{support\_\allowbreak{}contact} \\[3pt]
\texttt{place\_\allowbreak{}shoe} & S: \texttt{placed\_\allowbreak{}at\_\allowbreak{}fixed\_\allowbreak{}target\_\allowbreak{}oriented};\allowbreak{} F: \texttt{support\_\allowbreak{}contact} \\[3pt]
\texttt{put\_\allowbreak{}bottles\_\allowbreak{}dustbin} & S: \texttt{all\_\allowbreak{}bottles\_\allowbreak{}in\_\allowbreak{}fixed\_\allowbreak{}region};\allowbreak{} G: \texttt{handover\_\allowbreak{}receive} \\[3pt]
\texttt{put\_\allowbreak{}object\_\allowbreak{}cabinet} & S: \texttt{cabinet\_\allowbreak{}target\_\allowbreak{}region\_\allowbreak{}entry};\allowbreak{} S: \texttt{expert\_\allowbreak{}drawer\_\allowbreak{}pull};\allowbreak{} G: \texttt{handle\_\allowbreak{}grasp} \\[3pt]
\texttt{rotate\_\allowbreak{}qrcode} & S: \texttt{oriented\_\allowbreak{}below\_\allowbreak{}height} \\[3pt]
\texttt{scan\_\allowbreak{}object} & S: \texttt{scanner\_\allowbreak{}axis\_\allowbreak{}alignment} \\[3pt]
\texttt{shake\_\allowbreak{}bottle}, \texttt{shake\_\allowbreak{}bottle\_\allowbreak{}horizontally} & S: \texttt{lift\_\allowbreak{}height\_\allowbreak{}proxy} \\[3pt]
\texttt{stack\_\allowbreak{}blocks\_\allowbreak{}three} & S: \texttt{three\_\allowbreak{}block\_\allowbreak{}stack\_\allowbreak{}geometry};\allowbreak{} F: \texttt{stack\_\allowbreak{}contact} \\[3pt]
\texttt{stack\_\allowbreak{}blocks\_\allowbreak{}two} & S: \texttt{two\_\allowbreak{}block\_\allowbreak{}stack\_\allowbreak{}geometry};\allowbreak{} F: \texttt{stack\_\allowbreak{}contact} \\[3pt]
\texttt{stack\_\allowbreak{}bowls\_\allowbreak{}three} & S: \texttt{three\_\allowbreak{}bowl\_\allowbreak{}stack\_\allowbreak{}geometry};\allowbreak{} F: \texttt{stack\_\allowbreak{}contact} \\[3pt]
\texttt{stack\_\allowbreak{}bowls\_\allowbreak{}two} & S: \texttt{two\_\allowbreak{}bowl\_\allowbreak{}stack\_\allowbreak{}geometry};\allowbreak{} F: \texttt{stack\_\allowbreak{}contact} \\[3pt]
\texttt{stamp\_\allowbreak{}seal} & S: \texttt{seal\_\allowbreak{}at\_\allowbreak{}visual\_\allowbreak{}target\_\allowbreak{}xy} \\[3pt]
\texttt{turn\_\allowbreak{}switch} & S: \texttt{joint\_\allowbreak{}near\_\allowbreak{}upper\_\allowbreak{}limit} \\[3pt]
\bottomrule
\end{tabularx}
\end{table}